\pdfoutput=1
\documentclass[11pt]{article}

\usepackage{acl}

\usepackage{times}
\usepackage{latexsym}

\usepackage[T1]{fontenc}
\usepackage[utf8]{inputenc}

\usepackage{microtype}

\usepackage{graphicx}
\usepackage{booktabs}
\usepackage{multicol}

\usepackage{amsmath}
\usepackage{scalerel,amssymb}
\usepackage{comment}
\usepackage{adjustbox}
\def\mcirc{\mathbin{\scalerel*{\circ}{j}}}
\def\msquare{\mathord{\scalerel*{\Box}{gX}}}

\def\v#1{\mathbf{#1}}

\newcommand{\thickhline}{
    \specialrule{.1em}{0em}{0em}
}

\definecolor{Mahogany}{rgb}{0.75, 0.25, 0.0}

\newcommand{\gs}[1]{}
 \newcommand{\kl}[1]{}
 \newcommand{\bs}[1]{}

\newcommand{\kledit}[1]{{#1}}
\newcommand{\bsedit}[1]{{#1}}

\title{Searching for fingerspelled content in American Sign Language}
\author{Bowen Shi \\
  TTI-Chicago \\
  \texttt{bshi@ttic.edu} \\\And
  Diane Brentari \\
  Univeristy of Chicago \\
  \texttt{dbrentari@uchicago.edu}\\
  \AND
  Greg Shakhnarovich \\
  TTI-Chicago \\
  \texttt{greg@ttic.edu}\\\And  
  Karen Livescu \\
  TTI-Chicago \\
  \texttt{klivescu@ttic.edu}  
  \\}
  
\begin{document}
\maketitle
\begin{abstract}
% 
% 
% .
Natural language processing for sign language video---including tasks like recognition, translation, and search---is crucial for making artificial intelligence technologies accessible to deaf individuals, and is gaining research interest in recent years.  In this paper, we address the problem of searching for fingerspelled keywords or key phrases in raw sign language videos.  This is an important task since significant content in sign language is often conveyed via fingerspelling, and to our knowledge the task has not been studied before.  We propose an end-to-end model for this task, FSS-Net, that jointly detects fingerspelling and matches it to a text sequence. Our experiments, done on a large public dataset of ASL fingerspelling in the wild, show the importance of fingerspelling detection as a component of a search and retrieval model.  Our model significantly outperforms baseline methods adapted from prior work \kledit{on related tasks}. 

\end{abstract}

\section{Introduction}
\label{sec:intro}

Sign languages \kledit{are} a type of natural \kledit{language} which convey meaning through sequences of handshapes and gestures as well as non-manual elements, \kledit{and} are \kledit{a} chief means of communication for about 70 million deaf people \kledit{worldwide.}%
\footnote{From \texttt{https://wfdeaf.org/our-work/}} 
Automatic sign language 
\kledit{technologies} would \kledit{help to} bridge the \kledit{communication barrier} between deaf and hearing individuals\kledit{, and would make deaf video media more searchable and indexable.}

\begin{figure}[htp]
    \centering
    \includegraphics[width=\linewidth]{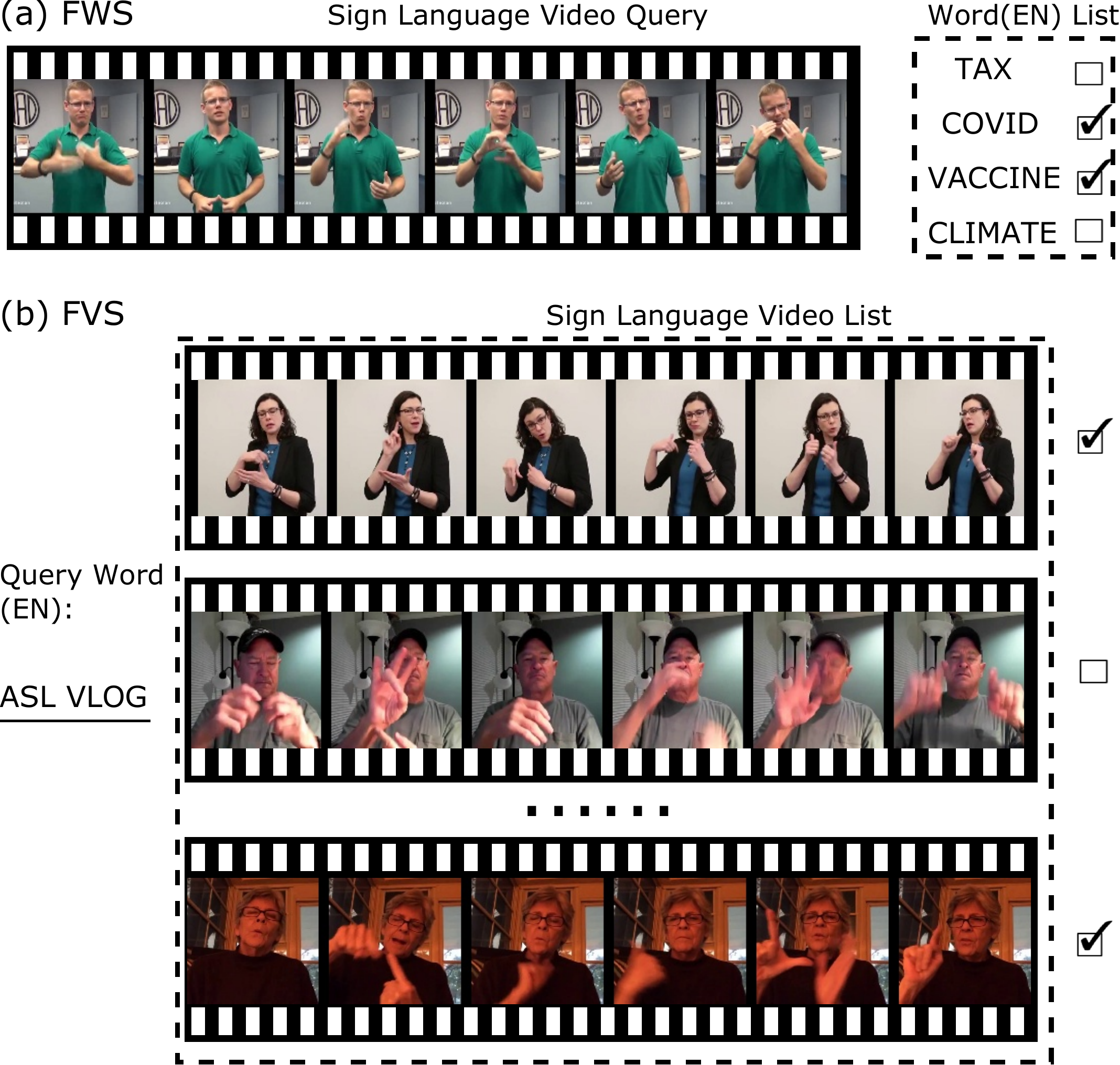}
       \caption{\label{fig:task}
       Our two search tasks:  (a) \emph{fingerspelled word search} (FWS) for determining which words are fingerspelled in a sign language video \kledit{clip},
       and (b) \emph{fingerspelling video search} (FVS) for searching for sign language videos that include a fingerspelled query word/phrase. The sign language \kledit{videos are untrimmed, i.e.~they include} regular signs in addition to fingerspelling\kledit{, and
       are downsampled here for visualization.}
       }
      \vspace{-0.0in}
\end{figure}

Automatic sign language 
\kledit{processing} has recently 
received growing interest in \kledit{the computer vision (CV) and natural language processing (NLP)} communities. 
\kledit{\citet{yin-etal-2021-including} make several recommendations for the study of sign languages in NLP research, including greater emphasis on real-world data.} 
Most studies \kledit{on sign language} are based on data collected \kledit{in} a controlled environment, either 
\kledit{in a studio setting} ~\citep{Martnez2002PurdueRA,Kim2017LexiconfreeFR} or in a specific domain~\citep{forster-etal-2014-extensions}. The challenges involved in \kledit{real-world} 
signing videos, including \kledit{various} visual conditions and different levels of fluency in signing, are \kledit{not} fully reflected in \kledit{such} datasets. Automatic \kledit{processing of sign language}
videos \kledit{"in the wild"} has not been addressed until recently, and is still restricted to % 
\kledit{tasks like} isolated sign recognition~\citep{albanie2020bsl1k,Joze2019MSASLAL,Li2020WordlevelDS} and fingerspelling recognition~\citep{shi2018slt,shi2019iccv}.  \kledit{In this work we take a step further and study search and retrieval of arbitrary fingerspelled content in real-world American Sign Language (ASL) video (see Figure~\ref{fig:task}).}  %

\kledit{Fingerspelling is a 
component} of sign language in which words are signed by a series of handshapes or movements corresponding to single letters \kledit{(see the Appendix for the ASL fingerspelling alphabet).}
Fingerspelling is \kledit{used mainly} for lexical items that do not have their own signs, such as proper nouns or technical terms, and has an important place in sign language. For example, fingerspelling accounts for 12-35\% of
ASL~\citep{padden2003alphabet}.  \kledit{Since important content like named entities is often fingerspelled,
the \kledit{fingerspelled portions} of a 
sign language video often carry a disproportionate amount of the content.} 
Most prior \kledit{work} on fingerspelling \kledit{has} focused on recognition~\citep{shi2018slt,shi2019iccv}, \kledit{that is, transcription of a fingerspelling} 
video clip into text. However, automatic recognition \kledit{assumes that the boundaries of fingerspelled segments are known at test time, and may not be the end goal in real-world use cases.  In addition, complete transcription may not be necessary to extract the needed information.} 
\kledit{Fingerspelling search,} such as retrieving sign language videos based on a query \kledit{word,} is a more practical task\kledit{, and is an important component of general video search involving sign language}. %

In addition to \kledit{introducing} the task, \kledit{we address the research question of}
whether the explicit temporal localization of fingerspelling can help its search and retrieval\kledit{, and how best to localize it}. 
As fingerspelling occurs sparsely in the signing stream, \kledit{explicit} detection of fingerspelling could potentially improve \kledit{search} performance by removing unrelated signs. 
To this end, we propose an end-to-end \kledit{model,} FSS-Net, which jointly detects fingerspelling from \kledit{unconstrained} signing video and 
\kledit{matches} it to \kledit{text queries}. Our approach consistently outperforms a series of baselines without explicit detection \bsedit{and a baseline with an off-the-shelf fingerspelling detector} by a large margin. % 

\section{Related Work}
\label{sec:related_work}
In existing \kledit{work on} sign language \kledit{video} processing, search and retrieval tasks \kledit{have been} studied much less than sign language recognition \kledit{(mapping from sign language video to gloss labels)}~\citep{koller2017resign,forster-etal-2014-extensions} and translation \kledit{(mapping from sign language video to text in another language)}~\citep{yin2020better,Camgoz2018NSLT}.
\kledit{Work thus far on sign language search has been framed mainly} as the retrieval of \bsedit{lexical signs \kledit{rather than} fingerspelling}.  %
\citet{Pfister2013LargescaleLO,albanie2020bsl1k} employ mouthing to detect keywords in sign-interpreted TV programs with coarsely aligned subtitles. \citet{Tamer2020CrossLingualKS,Tamer2020KeywordSF} utilize whole-body pose estimation to search \kledit{for} sign language keywords (gloss or translated word) \kledit{in a} German Sign Language translation dataset PHOENIX-2014T~\citep{Camgoz2018NSLT}. %
All prior work \kledit{on} keyword search for sign language \kledit{has been done in} a closed-vocabulary setting, which assumes that only words \bsedit{from a pre-determined set} will be queried. %
\kledit{Searching in an open-vocabulary setting, including proper nouns, typically requires searching for fingerspelling.}
\kledit{Some related tasks in the speech processing literature are spoken term detection (STD) and query-by-example search, which are the tasks of automatically retrieving speech segments 
from a 
database that match a given text or audio query~\cite{knill2013investigation,mamou2007vocabulary,Chen2015QuerybyexampleKS}}. 
In terms of methodology, our model \kledit{also shares some aspects with prior work on moment retrieval}~\citep{gao2017tall,Xu2019MultilevelLA,zhang2020span}, \kledit{which also combines candidate generation and matching components}.
\kledit{However, we incorporate additional}
\bsedit{task-specific}
\kledit{elements 
that consistently improve performance}. 

\section{Tasks}
\label{sec:task}
We consider two \kledit{tasks:}
Fingerspelled Word Search (FWS) and Fingerspelling-based Video Search (FVS). FWS and FVS respectively \kledit{consist} of detecting % 
fingerspelled \kledit{words} within a given raw ASL video stream and detecting \kledit{video clips} of interest containing a given fingerspelled word.\footnote{\kledit{We use "word" to refer to a fingerspelling sequence, which could be a single word or a phrase.}} % 
Given a query video clip $v$ and a list of $n$ words $w_{1:n}$, \kledit{FWS is the task of} finding 
\kledit{which (if any) of $w_{1:n}$ are present in $v$.} 
Conversely, \kledit{in FVS the input} is a query word $w$ and $n$ video clips $v_{1:n}$, and the task consists of finding 
all videos \kledit{containing the fingerspelled word $w$}. We consider \kledit{an} open-vocabulary setting where the word $w$ is \kledit{not constrained} to a pre-determined set.
\kledit{The two tasks correspond to} two directions of search (video$\longrightarrow$text and text$\longrightarrow$video), 
\kledit{as is standard practice in other retrieval work such as} video-text search~\citep{zhang2018crossmodal,krishna2017dense,ging2020coot}.

\section{Model}
\label{sec:models}

\begin{figure*}[htb]
  \centering
\adjustbox{max width=1\textwidth}{
  \includegraphics[width=1.0\linewidth]{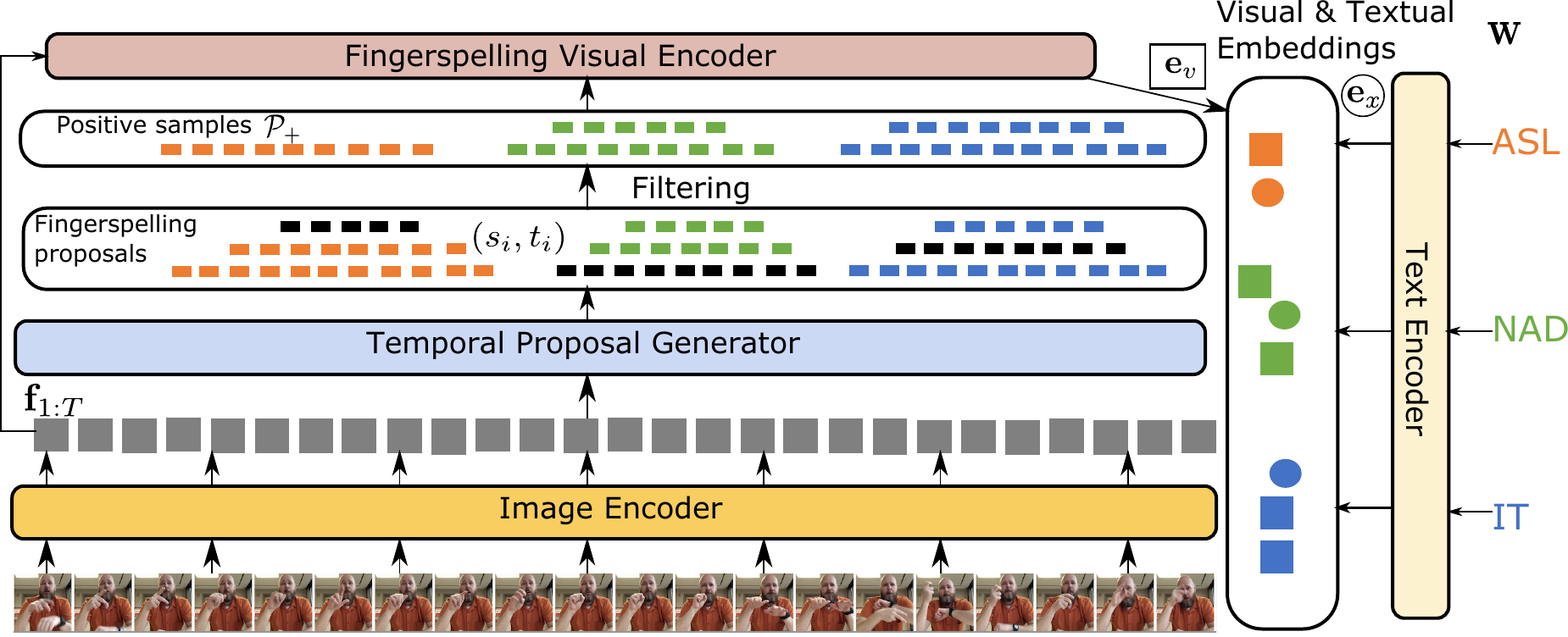}
  }
      \caption{\label{fig:fss-net-model}  FSS-Net: The proposed model for fingerspelling search and \kledit{retrieval.} The model maps candidate fingerspelling segments and text into a shared embedding space. $\mcirc$: text embedding, $\msquare$: visual embedding. The colors \kledit{correspond to  different input fingerspelling sequences}. \kledit{As pictured, this is the training time model, where the pairing between text and video segments is known.  At test time, the labels (colors) of the visual embeddings are unknown \bsedit{and we do not filter the proposals}. }}
      \vspace{-0.2in}
\end{figure*}

We propose a single \kledit{model, FSS-Net (for "FingerSpelling Search Network"),} summarized in \kledit{Figure}~\ref{fig:fss-net-model}, to address the two aforementioned search tasks. FSS-Net receives a pair of \kledit{inputs---a raw ASL video clip, and a written text sequence---and produces a score indicating the degree of match between the video clip and the text}. The text is encoded into \bsedit{an embedding} vector via a learned encoder. The visual branch of \kledit{FSS-Net} generates a number of fingerspelling segment proposals and each proposed visual segment is  encoded into a feature space shared with the text embeddings. Paired embeddings from both modalities are drawn towards each other in the shared embedding space during training.

\textbf{Image encoding} The input image frames are encoded into a sequence of feature vectors via \bsedit{an image} encoder, which consists of \kledit{the} VGG-19~\citep{Simonyan15vgg} convolutional layers followed by \kledit{a} Bi-LSTM.\footnote{Transformers~\citep{Vaswani17transformer} can also be used, \kledit{but in our initial experiments, they were outperformed by BiLSTMs on our tasks and data}.} 
We use raw RGB images as \kledit{input,} instead of signer pose as used in \kledit{some prior work}~\citep{Tamer2020KeywordSF,Tamer2020CrossLingualKS} on sign language search, as estimating pose for hands is particularly hard for signing videos in the wild (see \kledit{Section}~\ref{sec:exp} for details). 

\textbf{Temporal proposal generation} 
Suppose the visual feature sequence is $\v f_{1:T}$, where $T$ is the number of frames \kledit{in} the video clip.
The purpose of temporal proposal generation is to produce a number of candidate fingerspelling segments \bsedit{$\mathcal{H}(\v I_{1:T})=\{(s_i, t_i)\}_{1\leq i\leq |\mathcal{H}(\v I_{1:T})|}$ % 
from $\v f_{1:T}$, where $s_i$,$t_i$ \kledit{are} the start and end frame indices of \kledit{the $i^{\textrm{th}}$ proposed} segment. Below we use $\mathcal{H}$ as a shorthand for $\mathcal{H}(\v I_{1:T})$.} Here we adopt the \kledit{strategy} in~\citep{xu2017rc3d}, \bsedit{which is commonly used to generate proposals} for action detection. \bsedit{Briefly, the model assigns a probability $p_{det}$ of each proposal being fingerspelling.} \bsedit{See~\citep{xu2017rc3d} for more details. We denote the detection loss as $L_{det}$.} %

% \begin{comment}

% \end{comment}

Note \kledit{that} the training requires known ground-truth fingerspelling \kledit{boundaries}. \kledit{In the fingerspelling datasets we use here~\citep{shi2018slt,shi2019iccv}, the fingerspelling boundaries are already annotated, so no further annotation is needed.}

\textbf{Filtering} 
\kledit{A} visual embedding is produced 
\kledit{for each} segment.
\kledit{We} denote \bsedit{a labeled fingerspelling segment (shortened as fingerspelling segment below)} as a tuple $(s, t, w)$, where $s$, $t$ and $w$ represent \kledit{the} start frame index, \kledit{the} end frame index, and the written text it represents. 
\kledit{A} naive approach 
\kledit{would be to use only the} ground-truth fingerspelling segments $\mathcal{P}_g=\{(s_i, t_i, w_i)\}_{1\leq i\leq |\mathcal{P}_g|}$ \kledit{for training}.
However, \kledit{this} approach does not take \kledit{into account the potential shifts (errors) that may exist at test time between the ground-truth and generated segment proposals}. 
The \kledit{embeddings} produced by the \kledit{fingerspelling encoder} %
should be robust to such shifts. To this end, we incorporate proposals in forming positive pairs at training time. Formally, 
\kledit{let} the set of time intervals from \kledit{the} temporal proposal generator \kledit{be}  $\mathcal{H}=\{(s_i, t_i)\}_{1\leq i\leq |\mathcal{H}|}$. 
We sample $K$ intervals from $\mathcal{P}_t$ to form the set of generated fingerspelling segments: 
\begin{equation}
\vspace{-0.1in}
\label{eq-pos-pairs}
\begin{split}
\mathcal{P}_k = &\{(s_k, t_k, w_g)|IoU((s_k, t_k), (s_g, t_g))>\delta_{IoU}, \\
& IS((s_t, t_k), (s_g, t_g))>\delta_{IS}, \\
& (s_k, t_k)\in \mathcal{H}, (s_g, t_g, w_g)\in\mathcal{P}_g\}\\
\end{split}
\vspace{-0.08in}
\end{equation}
\noindent \kledit{where} 
$\text{IS}(x, y)=\frac{\text{Intersection}(x,y)}{\text{Length}(y)}$ and $\text{IoU}(x, y)=\frac{\text{Intersection}(x,y)}{\text{Union}(x,y)}$. We use $\delta_{IoU}$ and $\delta_{IS}$ to control the degree to which the proposals can deviate from the ground-truth.  In addition to \kledit{the intersection over union (IoU)}, we use \kledit{the normalized intersection IS} to eliminate proposals with many missing frames.
We take the union of the two \kledit{sets, $\mathcal{P}_{+}=\mathcal{P}_g \cup\mathcal{P}_k$, as the filtered proposal set to be encoded.} 

 \textbf{Fingerspelling visual encoding \kledit{(FS-encoding)}}
\bsedit{The visual encoding of each segment $(s, t,  w)\in\mathcal{P}_+$ is $\v e_{v}^{(s,t)}=\text{Bi-LSTM}(\v f_{s: t})$.}\footnote{We compared the Bi-LSTM encoder with average/max pooling of $f_{s:t}$, and found the former to perform better.}

\textbf{Text encoding} A written word (or phrase) $\v w$ is mapped to an embedding vector $\v e_x^{w}$ via a text encoder. To handle words \kledit{not seen at training time (and better handle rarely seen words)}, we first decompose $\v w$ into a sequence of characters $c_{1:|w|}$ (e.g. `ASL'=`A'-`S'-`L') and feed the character sequence $c_{1:|w|}$ into a text encoder (\kledit{here, a} Bi-LSTM\footnote{\kledit{Again, transformers can also be used, but in our experiments Bi-LSTMs show} better performance.}). 

\textbf{Visual-text matching} With the above pairs of visual and textual embeddings, we use a training objective function consisting of two triplet loss terms: 
\begin{equation}
\centering
\label{eq:fss-net-triplet}
\begin{split}
& L_{tri}(\v I_{1:T}, \mathcal{P}_+) =\\ &\displaystyle\sum_{(s,t, w)\in\mathcal{P}_{+}}\max\{m+d(\v e^{(s,t)}_v, \v e^{w}_x)\\ 
& -\frac{1}{|\mathcal{N}_w|}\displaystyle\sum_{ w^\prime\in\mathcal{N}_w}d(\v e^{(s,t)}_v, \v e_x^{w^\prime}), 0\} \\
& + \max\{m+d(\v e^{(s,t)}_v, \v e^w_x)\\
& -\frac{1}{|\mathcal{N}_v|}\displaystyle\sum_{(s^\prime,t^\prime)\in\mathcal{N}_v}d(\v e^{(s^\prime,t^\prime)}_v, \v e^w_x), 0\} \\
\end{split}
\end{equation}
\noindent where \kledit{$d$} denotes cosine distance $d(\v a,\v b)=1-\frac{\v a\cdot\v b}{\|\v a\|\|\v b\|}$\kledit{, $m$ is a margin, and}
%.
$\mathcal{N}_v$ and $\mathcal{N}_{w}$ are \kledit{sets} of negative samples of proposals and words. To form negative pairs we 
\kledit{use} semi-hard negative \kledit{sampling~\citep{Schroff2015FaceNetAU}:}
\begin{equation}
\label{eq:semi-hard-neg}
\centering
\begin{split}
\mathcal{N}_v &= \{(s^\prime, t^\prime)|d(\v e^{(s^\prime, t^\prime)}_v, \v e^w_x) > d(\v e^{(s,t)}_v, \v e^w_x)\} \\
\mathcal{N}_w &= \{w^\prime|d(\v e^{(s,t)}_{v}, \v e^{w^\prime}_x) > d(\v e^{(s,t)}_v, \v e^w_x)\} \\
\end{split}
\end{equation}
\noindent For efficiency, negative samples are selected 
\kledit{from the corresponding} mini-batch. 

\textbf{Overall loss} The model is trained %jointly 
with \kledit{a combination of the} detection loss and triplet loss: %
\begin{equation}
\label{eq:fss-overall-loss}
\resizebox{\linewidth}{!} 
{
$ L_{tot}(\v I_{1:T}, \mathcal{P}_g) = \lambda_{det} L_{det}(\v I_{1:T}, \mathcal{P}_g) + L_{tri}(\v I_{1:T}, \mathcal{P}_+) $
}
\end{equation}
 with \kledit{the} tuned weight  $\lambda_{det}$ controlling the relative importance of detection versus visual-textual matching.%

\textbf{Inference} At test time, the model assigns a score $sc(\v I_{1:T}, w)$ \kledit{to a given video clip $\v I_{1:T}$ and word $w$.} The word is encoded into the word embedding $\v e^{w}_x$.
\bsedit{Suppose the set of fingerspelling proposals generated by the temporal proposal generator is $\mathcal{H}(\v I_{1:T})$.}
% 

% 
% % .
% 
% 
%
% 
%
%
%
\kledit{We define a} scoring function for the proposal $h\in\mathcal{H}(\v I_{1:T})$ and word $w$ 
\begin{equation}
\label{eq:fss-search-score}
{sc}_{word}(h_m, w)=p_{det}^\beta(1-d(\v e_v^{h_m}, \v e_x^w))
\end{equation}
\noindent where $p_{det}$ is the probability  \bsedit{given by the temporal proposal generator}
and $\beta$ controls the relative weight between detection and matching.  \kledit{In other words, in order for a segment and word to receive a high score, the segment should be likely to be fingerspelling (according to $p_{det}$) and its embedding should match the text.} 
 Finally, the \kledit{score for the} video clip $\v I_{1:T}$ and the word $w$ 
 \kledit{is defined as} the highest score among \bsedit{the set of proposals $\mathcal{H}(\v I_{1:T})$:}
\begin{equation}
\label{eq:fss-final-score}
{sc}(\v I_{1:T},w)=\displaystyle\max_{\begin{subarray}{l}{h\in\mathcal{H}(\v I_{1:T})}\end{subarray}}{sc}_{word}(h, w)
\end{equation}
\section{Experimental Setup}
\subsection{Data}

We conduct experiments on ChicagoFSWild~\citep{shi2018slt} and ChicagoFSWild+~\citep{shi2019iccv}, two large-scale publicly available fingerspelling datasets containing 7,304 and 55,272 fingerspelling sequences respectively. The ASL videos in the two datasets are collected from online resources and include a variety of viewpoints and styles, such as webcam videos and lectures. 

We follow the setup of~\citep{shi2021cvpr} \kledit{and} split \kledit{the} raw ASL videos into 300-frame clips with a 75-frame overlap between neighboring chunks and remove clips without fingerspelling. 
The \kledit{numbers} of clips in \kledit{the various splits}
\bsedit{ can be found in the Appendix.}
On average, each clip contains 1.9/1.8 fingerspelling segments in the ChicagoFSWild and ChicagoFSWild+ datasets respectively.

\subsection{Baselines}
\kledit{We compare the proposed model, FSS-Net,} % 
to the following baselines adapted from common approaches for search and retrieval in related fields.  % 
To facilitate comparison, the network architecture for \kledit{the} visual and text encoding in \kledit{all baselines is the same as in} FSS-Net. 

\textbf{Recognizer} In this approach, we train a % 
\kledit{recognizer that transcribes the video clip into text.}
\kledit{Specifically, we train a recognizer to output a sequence of symbols consisting of either fingerspelled letters or a special non-fingerspelling symbol <x>.} 
\kledit{We train the recognizer with a connectionist temporal classification (CTC) loss~\citep{ctc}, which is commonly used for speech recognition}. At test time, we use beam search to generate a list of hypotheses $\hat{\v w}_{1:M}$ for the target video clip $\v I_{1:T}$. Each hypothesis $\hat{w}_m$ is split into a list of words $\{\hat{w}^n_m\}_{1\leq n\leq N}$ \kledit{separated} by <x>. The matching score between video $\v I_{1:T}$ and \kledit{$\v w$} is defined as:
\begin{equation}
    \label{eq:word-list-score}
    {sc}(\v I_{1:T}, w)=1-\displaystyle\min_{1\leq m\leq M}\displaystyle\min_{1\leq n\leq N}\text{LER}(\hat{w}^n_m, w)
\end{equation}
\noindent where \kledit{the letter error rate} $\text{LER}$ \kledit{is the Levenshtein edit distance.}
%. 
% . 
This approach is adapted from~\citep{Saralar2004LatticeBasedSF} for spoken utterance retrieval. \bsedit{Fingerspelling boundary information is not used in \kledit{training} this baseline \kledit{model}.}

\textbf{Whole-clip} \kledit{The} whole-clip baseline encodes the whole video clip $\v I_{1:T}$ into a visual embedding  $\v e_v^{I}$, which \kledit{is} matched to the textual embedding $\v e_{x}^w$ of \kledit{the query} $\v w$. 
The model is trained with contrastive loss as in equation~\ref{eq:fss-net-triplet}. At test time, the 
\kledit{score for video clip $\v I_{1:T}$ and word $\v w$ is}:
\begin{equation}
    \label{eq:word-list-score}
    {sc}(\v I_{1:T},\v w)=1-d(\v e_{v}^{I}, \v e_{x}^{w})
\end{equation}
\noindent where $d$ is the cosine distance \kledit{as} in FSS-Net. \bsedit{Fingerspelling boundary information is \kledit{again} not used in this baseline.}

%%  

% .

\textbf{External detector (Ext-Det)} \kledit{This baseline uses} 
the off-the-shelf fingerspelling detectors of ~\citep{shi2021cvpr} to generate fingerspelling proposals\kledit{, instead of our proposal generator, and is otherwise identical to FSS-Net.} \kledit{For each dataset (ChicagoFSWild, ChicagoFSWild+), we use the detector trained on the training subset of that dataset.} This baseline uses ground-truth fingerspelling boundaries for the detector training.

\textbf{Attention-based \kledit{keyword search} (Attn-KWS)} % 
\kledit{This model is} 
adapted from~\citep{Tamer2020KeywordSF}\kledit{'s approach for} keyword search in sign language. The model employs \kledit{an} attention mechanism to match \kledit{a} text query with \kledit{a} video clip, where each frame is weighted based on the query embedding.  The attention mechanism enables the model to implicitly localize frames relevant to the text. \kledit{The model of~\citep{Tamer2020KeywordSF} is designed for lexical signs rather than fingerspelling.} To adapt the model \kledit{to our} open-vocabulary fingerspelling setting, we use the same text encoder as \kledit{in} FSS-Net to map words into embeddings instead of using \kledit{a} word embedding matrix as in~\citep{Tamer2020KeywordSF}. \kledit{Fingerspelling boundary information is again not used in training this model, which arguably puts it at a disadvantage compared to FSS-Net.} More details on the formulation of the model can be found in the \kledit{Appendix}.

 \subsection{Evaluation}
 For FWS, we use all words in the test set as the test vocabulary \kledit{$w_{1:n}$}. For FVS, all video clips in the test are used as candidates \kledit{and the text queries are again the entire test vocabulary}. 
 \bsedit{We report the results in terms of standard metrics from the video-text retrieval literature~\citep{Momeni2020SeeingWW,Tamer2020CrossLingualKS}: mean Average Precision (mAP)} \kledit{and mean F1 score (mF1), where the averages are over words for FVS and over videos for FWS.}
 Hyperparameters are chosen to maximize the mAP on the dev set, independently for the two tasks (though ultimately, the best hyperparameter values in our search are identical for both tasks).  \kledit{Additional details} on data, preprocessing, model implementation, and hyperparameters can be found in the Appendix. 
\section{Results and analysis}
\label{sec:exp}

\subsection{Main Results}
\begin{table}[htb]
    \caption{\label{tab:fss-perf-fswild}FWS/FVS 
    \kledit{performance} on \kledit{the} ChicagoFSWild and ChicagoFSWild+ \kledit{test sets}. \bsedit{The range of mAP and mF1 is [0, 1].}
    $\bigstar$: methods that use fingerspelling boundaries in training.}
    % \vspace{-0.1in}
    % \vspace{5pt}
    \centering
    % \footnotesize
        \setlength{\tabcolsep}{3pt}

    \resizebox{\columnwidth}{!}{
    \begin{tabular}{l c c c c }
        \thickhline
        \multicolumn{1}{c}{} &
        \multicolumn{2}{c}{FWS (Video$\implies$Text)} &
        \multicolumn{2}{c}{FVS (Text$\implies$Video)}   \\
        \cmidrule(r){2-3}
        \cmidrule(r){4-5}
         \multicolumn{5}{c}{\textbf{ChicagoFSWild}} \\ 
        \cmidrule(r){2-3}
        \cmidrule(r){4-5}
        Method & mAP & mF1 & mAP & mF1 \\
        \cmidrule(r){2-3}
        \cmidrule(r){4-5}
        Whole-clip & .175 & .154 & .142 & .119  \\
        Attn-KWS & .204 & .181 & .246 & .229 \\
        Recognizer & .318 & .315 & .331 & .305  \\ 
        Ext-Det$\bigstar$ & .383 & .385 & .332 & .312 \\
        FSS-Net$\bigstar$ &\textbf{.434}&\textbf{.439} & \textbf{.394} & \textbf{.370}   \\
        \cmidrule(r){2-3}
        \cmidrule(r){4-5}
         \multicolumn{5}{c}{\textbf{ChicagoFSWild+}} \\ 
        \cmidrule(r){2-3}
        \cmidrule(r){4-5}
        Method & mAP & mF1 & mAP & mF1  \\
        \cmidrule(r){2-3}\cmidrule(r){4-5}
        Whole-clip & .466 & .457
        & .548 & .526 \\
        Attn-KWS & .545 & .530 &
        .573 & .547 \\
        Recognizer & .465 & .462 & .398 & .405  \\
        Ext-Det$\bigstar$ &
        .633 & .641 &
        .593 & .577 \\
        FSS-Net$\bigstar$ & \textbf{.674}
 & \textbf{.677} & \textbf{.638} &
\textbf{.631} \\ \thickhline
    \end{tabular}
    }
    \vspace{-0.15in}
\end{table}

 Table~\ref{tab:fss-perf-fswild} shows the performance of 
 \kledit{the} above approaches on \kledit{the} two datasets.  First, we \kledit{notice} that embedding-based approaches consistently outperform the recognizer baseline in the larger data setting (ChicagoFSWild+) \kledit{but not the smaller data setting (ChicagoFSWild), which %mainly 
 suggests that} embedding-based models generally require more training data. \kledit{The inferior performance of recognizer} also shows that \kledit{explicit} fingerspelling recognition
 %, 
 is not necessary for the \kledit{search tasks.} 
 In addition, explicit \kledit{fingerspelling detection (Ext-Det, FSS-Net)
 improves performance over} implicit fingerspelling detection (Attn-KWS) and detection-free \kledit{search} (Whole-clip).  \kledit{Explicit fingerspelling detection requires boundary information during training.  Of the models that don't use such supervision, Attn-KWS is the best performer given enough data, but is still far behind FSS-Net.} %
 Our model outperforms all \kledit{of the alternatives}. %
 The \kledit{relative performance of} different models remains consistent across \kledit{the} various metrics \kledit{and the} two search tasks. 
\bsedit{For completeness, we also measure the performance of different models in \kledit{terms of} ranking-based metrics (e.g., Precision@N, Recall@N), as in prior \kledit{work on} video-text retrieval~\citep{ging2020coot,krishna2017dense} \kledit{(see full results in the Appendix)}. The \kledit{relative} performance of different models \kledit{remains} consistent \kledit{on} these metrics.}
The analysis below is done on ChicagoFSWild for simplicity. The conclusions also hold for ChicagoFSWild+.
\newcommand\widthratio{0.99}
\begin{figure*}[htb]
\caption{\label{fig:kws-examples} Examples of FWS \kledit{predictions}. \kledit{For each example video, the ground truth (GT) is shown along with the top 5} predicted \kledit{fingerspelling sequences}. Top red line: ground-truth fingerspelling segment. Bottom blue line: \kledit{highest-scoring} predicted fingerspelling segment. \kledit{Segment locations are shown here for qualitative analysis, but they are not part of the task evaluation.}  \kledit{Note that many fingerspelling sequences (both ground-truth and predictions) are abbreviations, and some are misspelled; we include all fingerspelling sequences that appear in the \bsedit{test set} in the query vocabulary.} 
}
\vspace{-0.1in}
\begin{tabular}{l}
\toprule
\multicolumn{1}{c}{Successful retrieval}\\
\midrule
\midrule
GT: \textcolor{red}{LITERACYS TO} $\leftrightarrow
$ Pred: \textcolor{blue}{LITERACYS TO},  LITERACY, DISTRACT, ILOW, LIST  \\
\includegraphics[width=\widthratio\linewidth]{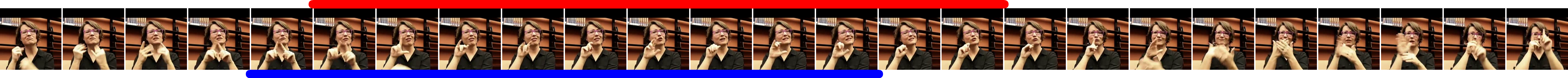}\\
\midrule
GT: \textcolor{red}{ASL} $\leftrightarrow
$ Pred: \textcolor{blue}{ASL}, ALL, ASLIED, ALLAH, HOME  \\
\includegraphics[width=\widthratio\linewidth]{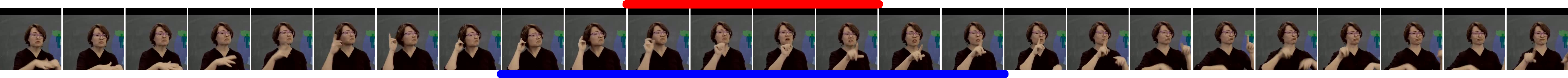}\\
\midrule
GT: \textcolor{red}{US} $\leftrightarrow
$ 
Pred: \textcolor{blue}{US}, USA, CAMUS, LS, SUCH AS  \\
\includegraphics[width=\widthratio\linewidth]{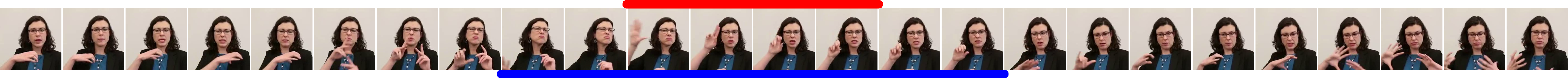}\\
\midrule
\multicolumn{1}{c}{Failure cases}\\
\midrule
\midrule
GT: \textcolor{red}{BACK} $\leftrightarrow
$ 
Pred: BA, AEBSP, BAK, AS, AT BTH BEACH  \\
\includegraphics[width=\widthratio\linewidth]{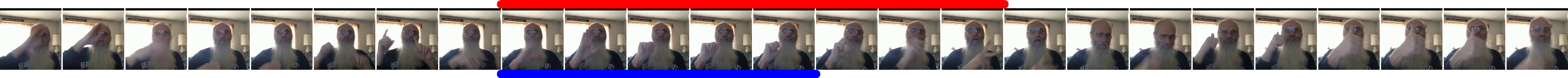}\\
\midrule
GT: \textcolor{red}{JETS} $\leftrightarrow
$ 
Pred: IT, OF, OFF, IE, IX  \\
\includegraphics[width=\widthratio\linewidth]{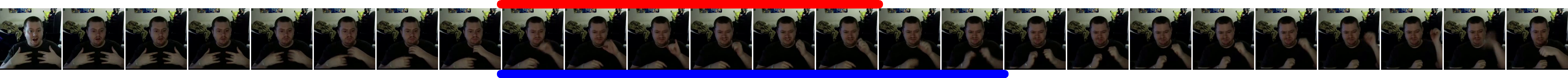}\\
\midrule
GT: \textcolor{red}{TXPU} $\leftrightarrow
$ 
Pred: FISH, F EST, RG, GER, TOSS  \\
\includegraphics[width=\widthratio\linewidth]{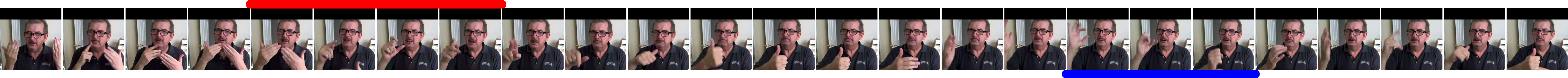}\\
\end{tabular}
\vspace{-0.2in}
\end{figure*}

\subsection{Model analysis}
\textbf{Does better localization lead to better \kledit{search}?}
\kledit{In the previous section we have seen that models that explicitly detect and localize fingerspelling outperform ones that do not.  Next we look more closely at how well several models---Ext-Det, Attn-KWS and FSS-Net---perform
on the task of localizing fingerspelling, which is a byproduct of these models' output.
We measure performance via AP@IoU, a commonly used evaluation metric for action detection~\citep{Haroon2016thumos,Heilbron2015ActivityNetAL} that has also been used for fingerspelling detection~\citep{shi2021cvpr}.  AP@IoU measures the average precision of a detector under the constraint that the overlap of its predicted segments with the ground truth is above some threshold Intersection-over-Union (IoU) value.
For Attn-KWS, the model outputs an attention vector, which we convert to segments as in~\citep{shi2021cvpr}. }

\begin{table}[htp]
    \vspace{-0.05in}
    \centering
        \caption{\label{tab:ap_iou}\kledit{Fingerspelling localization performance for detection-based models.}}
    \begin{tabular}{cccc}
    \toprule
         & AP@0.1 & AP@0.3 & AP@0.5 \\ 
         \midrule
        Attn-KWS & 0.268 & 0.104 & 0.035 \\
        Ext-Det & 0.495 & 0.453 & 0.344 \\
        Ours & \textbf{0.568} & \textbf{0.519} & \textbf{0.414} \\
        \bottomrule
    \end{tabular}
    \vspace{-0.05in}
\end{table}

In general, \kledit{the models with} more accurate localization 
\kledit{also have} higher search and retrieval performance\kledit{, as seen by comparing Table~\ref{tab:ap_iou} with Table~\ref{tab:fss-perf-fswild}}. 
\kledit{However, differences in AP@IoU do not directly translate to differences in search performance.}
For example, \kledit{the} AP@IoU of Ext-Det (0.344) is \kledit{an order of} magnitude higher than \kledit{that of} Attn-KWS (0.035) while their FVS mAP \kledit{results are much closer (0.593 vs.~0.573)}. %

\textbf{\kledit{Raw images vs.~estimated pose as input}} % 
Prior \kledit{work on sign language search~\citep{Tamer2020CrossLingualKS,Tamer2020KeywordSF} has used estimated pose keypoints as input, rather than raw images as we do here.} 
For comparison, we extract body and hand keypoints with \kledit{OpenPose}~\citep{openpose} and train a model with \kledit{the} pose \bsedit{skeleton} \kledit{as input}. 

\begin{table}[hbt]
    \caption{\label{tab:pose-vs-rgb}Impact of input \kledit{type (pose vs.~raw RGB images) on search} performance.}
    \vspace{-0.1in}
    % \vspace{5pt}
    \centering
    \setlength{\tabcolsep}{3pt}
    % \footnotesize
    \resizebox{\columnwidth}{!}{
    \begin{tabular}{l c c c c }
        \thickhline
        \multicolumn{1}{c}{} &
        \multicolumn{2}{c}{FWS (Video$\implies$Text)} &
        \multicolumn{2}{c}{FVS (Text$\implies$Video)}   \\
        \cmidrule(r){2-3}
        \cmidrule(r){4-5}
        Input & mAP & mF1 & mAP & mF1 \\ \cmidrule(r){2-3}\cmidrule(r){4-5}
        Pose & .142 & .147 & .127 & .121 \\
        RGB & \textbf{.434}&\textbf{.439} & \textbf{.394} & \textbf{.370}  \\
        \bottomrule
    \end{tabular}
    }
    \vspace{-0.1in}
\end{table}

As is shown in \kledit{Table}~\ref{tab:pose-vs-rgb}, \kledit{the} pose-based model 
\kledit{has much poorer search performance 
than the RGB image-based} models. \kledit{We believe this is largely because, while pose estimation works well for large motions and clean visual conditions, in our dataset much of the handshape information is lost
in the estimated pose (see the Appendix for some qualitative examples).} 

\subsection{Ablation Study} 

\begin{table}[hbt]
    \caption{\label{tab:full-ablation-study}Effect of \kledit{various} components of FSS-Net on FWS and \kledit{FVS.} 
    }
        \vspace{-0.1in}
    % \vspace{5pt}
    \centering
    \setlength{\tabcolsep}{3pt}
    % \footnotesize
    \resizebox{\columnwidth}{!}{
    \begin{tabular}{l c c c c }
        \thickhline
        \multicolumn{1}{c}{} &
        \multicolumn{2}{c}{FWS} &
        \multicolumn{2}{c}{FVS}   \\
        \cmidrule(r){2-3}
        \cmidrule(r){4-5}
         & mAP & mF1 & mAP & mF1 \\
        \cmidrule(r){2-3}
        \cmidrule(r){4-5}
        Full model & \textbf{.434}&\textbf{.439} & \textbf{.394} & \textbf{.370} \\
         \cmidrule(r){1-3}
         \cmidrule(r){4-5}
        (1) w/o generator & .186 & .180 & .259 & .270 \\
        (2) $\lambda_{det}=0$, $\beta=0$ & .411 & .420 & .373 & .350 \\
        (3) $\lambda_{det}=0.1$, $\beta=0$ & 
        .418 & .432 & .360 & .348  \\
        (4) w/o $\mathcal{P}_k$ & 
        .411 & .420 & .386 & .366 \\
        \bottomrule
    \end{tabular}
    }
        \vspace{-0.2in}
\end{table}

 Within our model, the proposal generator 
\kledit{produces a subset of all possible fingerspelling proposals, intended to represent the most likely fingerspelling segments.}
\kledit{To measure whether this component is important to the performance of the model, we compare our full model with the proposal generator to one where the proposal generator is removed (see Table~\ref{tab:full-ablation-study}).}
When the proposal generator is not used, the model is \kledit{trained only} with ground-truth fingerspelling segments ($\mathcal{P}_g$) and \kledit{considers all possible proposals within a set of sliding windows.}  
%. 
%
% 
\kledit{Such a "sliding-window" approach
is commonly used in previous} sign language keyword search~\citep{albanie2020bsl1k,Pfister2013LargescaleLO} \kledit{and} spoken \kledit{keyword spotting}~\citep{Chen2015QuerybyexampleKS}. As can be seen from \kledit{Table}~\ref{tab:full-ablation-study} \kledit{(Full model vs. row (1))}, the proposal generator greatly improves \kledit{search}
performance.  \kledit{This is not surprising, since the proposal generator greatly reduces the number of non-fingerspelling segments, thus lowering the chance of a mismatch between the text and video, and also refines the segment boundaries through regression, which should improve the quality of the visual segment encoding.}

\kledit{The fingerspelling detection component of our model has two aspects that may affect performance: imposing an additional loss during training,} and rescoring during inference. We disentangle \kledit{these} two factors and show their respective benefits \kledit{for} our model in \kledit{Table}~\ref{tab:full-ablation-study} \kledit{(row (2) and (3))}. The auxiliary detection task, which includes classification between fingerspelling and non-fingerspelling proposals, helps encode 
more comprehensive visual information 
into the visual embedding. 
In addition, the proposal probability output by the detector contains extra information and merging it into the matching score further improves the \kledit{search} performance.

Table~\ref{tab:full-ablation-study} \kledit{(row (4))} shows the effect of sampling additional proposals ($\mathcal{P}_k$) in fingerspelling detection. Additional positive samples make the visual embedding more robust to \kledit{temporal shifts} in the generated proposals, thus improving \kledit{search} performance. 

\subsection{Result analysis} 
\begin{figure}[htb]
\centering
    \vspace{-0.1in}
    \caption{\label{fig:performance-length}Performance as a function of fingerspelled word length. Red: FVS mAP, Blue: detection AP@IoU. }
    \includegraphics[width=\linewidth]{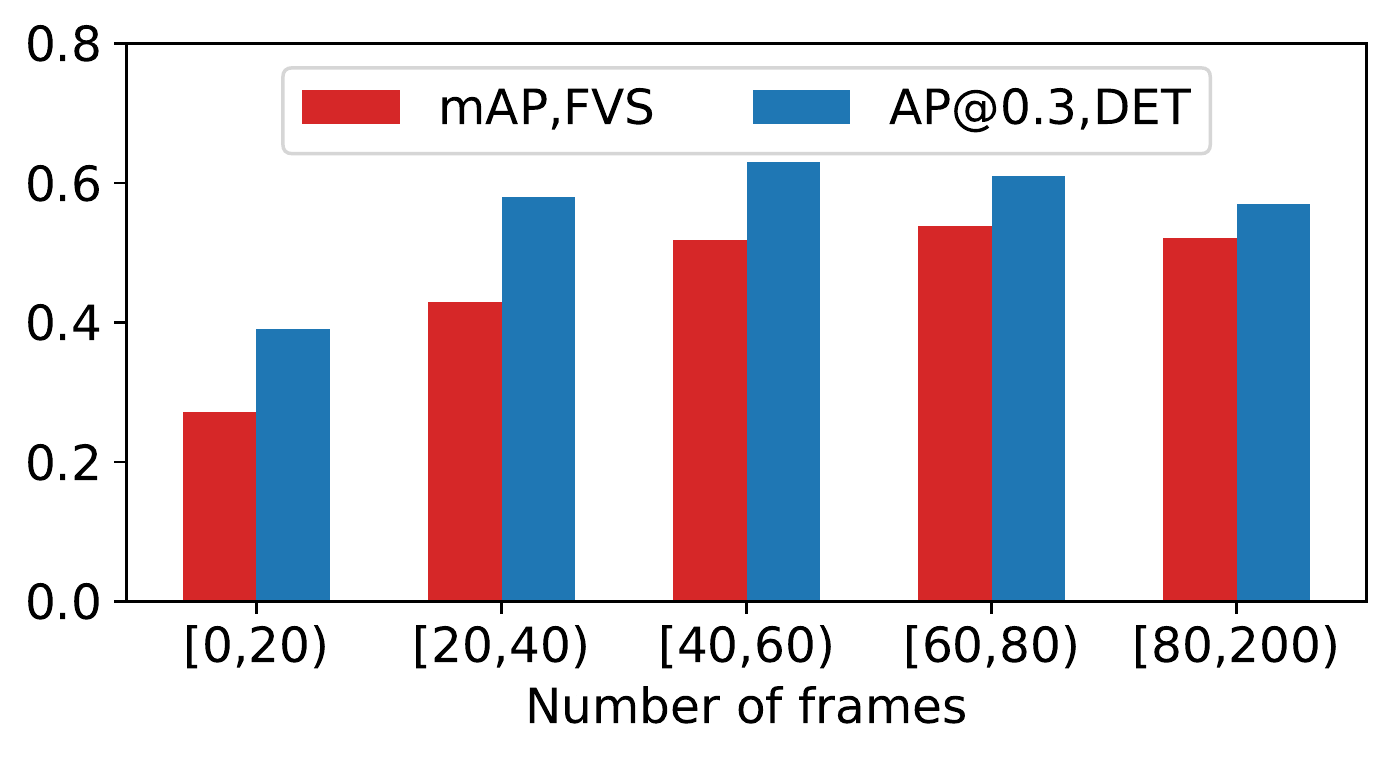}
    \vspace{-0.3in}
\end{figure}

The performance of our model is \kledit{worse} for short fingerspelled \bsedit{sequences} than for long \bsedit{sequences} (see \kledit{Figure}~\ref{fig:performance-length}). 
This 
\kledit{may be} because shorter words are harder to spot, as is shown from the trend in 
fingerspelling detection in the same figure.

The datasets we use are collected from multiple sources, and the video quality varies between them.  To quantify the effect of visual quality on search/retrieval performance, we categorize the ASL videos into three categories according to \kledit{their source:}
YouTube, DeafVIDEO, 
\kledit{and other miscellaneous sources (misc)}. YouTube videos are mostly ASL lectures with high resolution. DeafVIDEO videos are 
\kledit{vlogs} from deaf users of the social media \kledit{site} \texttt{deafvideo.tv}, where the style, \kledit{camera angle},
and image quality \kledit{vary greatly.}
The visual quality of \kledit{videos in the miscellaneous category tends to fall between the other} 
two categories. Typical image examples from the three categories can be found in the Appendix (figure~\ref{fig:source-example}).
\bsedit{The \kledit{FWS performance of our model on videos} in YouTube, DeafVIDEO, and misc are 0.684, 0.584, 0.629 (mAP) respectively.}
The results are overall consistent with the \kledit{perceived relative visual qualities of these categories}.

\kledit{As a qualitative analysis, we examine example words and videos on which our model is more or less successful.  Table~\ref{tab:map-examples} shows the query words/phrases with the highest/lowest FVS performance.  The best-performing queries tend to be long and drawn from the highest-quality video source.}

\begin{table}[htp]
    \centering
    \caption{\label{tab:map-examples}Example words with low/high mAP in FVS. The source of the corresponding video is given in parentheses}
        \vspace{-0.1in}
    \resizebox{\linewidth}{!}{
    \begin{tabular}{c|c}
    \toprule
         Low & High \\ 
         \midrule
         \begin{tabular}{@{}c@{}}
script (YouTube)\\
agent (misc)\\
kc (YouTube)\\
pati (DeafVIDEO)\\
mexer (DeafVIDEO)\\
flow (YouTube)\\
yr (DeafVIDEO)\\
exalted (misc)\\
poem (YouTube)\\
         \end{tabular}
 &          \begin{tabular}{@{}c@{}}
cabol erting (YouTube)\\
vp ron stern (YouTube)\\
co chairs (YouTube)\\
dr kristin mulrooney (YouTube)\\
myles (YouTube)\\
camaspace (YouTube)\\
electronics (YouTube)\\
brain (YouTube)\\
land (DeafVIDEO)\\
         \end{tabular} \\
        \bottomrule
    \end{tabular}}
        \vspace{-0.1in}
\end{table}

We also visualize the top \kledit{FWS} predictions made by our model in 
\kledit{several} video clips \kledit{(see figure~\ref{fig:kws-examples})}.
\kledit{Another common source of error is confusion between letters with similar handshapes (e.g., "i" vs. "j").  A final failure type is} 
fingerspelling detection failure. \kledit{As our model includes a fingerspelling detector, detection errors can harm search performance.}

\section{Conclusion}
\label{sec:conclusion}

Our work takes one step toward better addressing the need for language technologies for sign languages, by defining fingerspelling search tasks and developing a model tailored for these tasks.  These tasks are complementary to existing work on keyword search for lexical signs, in that it addresses the need to search for a variety of important content that tends to be fingerspelled, like named entities.  Fingerspelling search is also more challenging in that it requires the ability to handle an open vocabulary and arbitrary-length queries.  Our results demonstrate that a model tailored for the task in fact improves over baseline models based on related work on signed keyword search, fingerspelling detection, and speech recognition.  \kledit{However, there is room for improvement between our results and the maximum possible performance.  
One important aspect of our approach is the use of explicit fingerspelling detection within the model.  An interesting avenue for future work is to address the case where the training data does not include segment boundaries for detector training.  Finally, a complete sign language search system should consider both fingerspelling and lexical sign search.} 

\bibliography{anthology,custom}
\bibliographystyle{acl_natbib}

\clearpage
\appendix
\section{Appendix}
\label{sec:appendix}

\subsection{Fingerspelling alphabet}
\begin{figure}[htp]
    \centering
    \caption{\label{fig:alphabet}The ASL fingerspelling alphabet, from~\citep{jkean2014thesis}}
    \includegraphics[width=\linewidth]{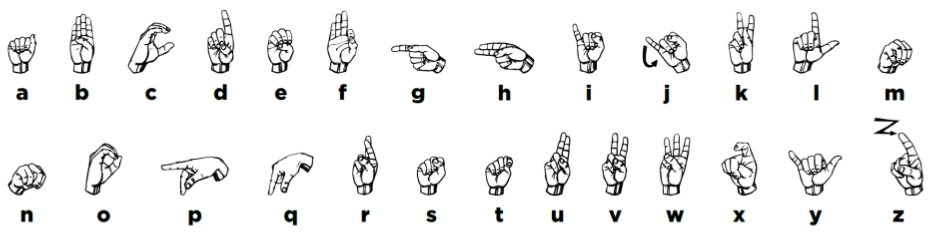}
\end{figure}

\subsection{Data}
\label{sec:app-data}
\begin{figure*}[htp]
\caption{\label{fig:source-example}Example image frames from different sources in ChicagoFSWild and ChicagoFSWild+.}
\begin{tabular}{ll}
\toprule
    YouTube & \raisebox{-.3\height}{\includegraphics[width=0.8\linewidth]{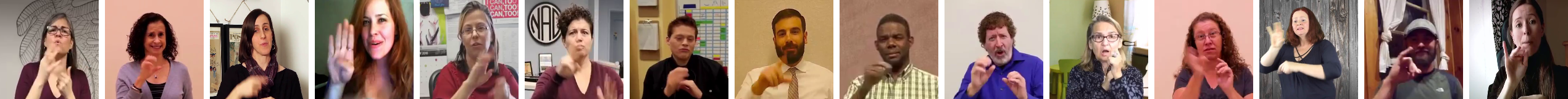}} \\
\midrule
    DeafVIDEO & \raisebox{-.3\height}{\includegraphics[width=0.8\linewidth]{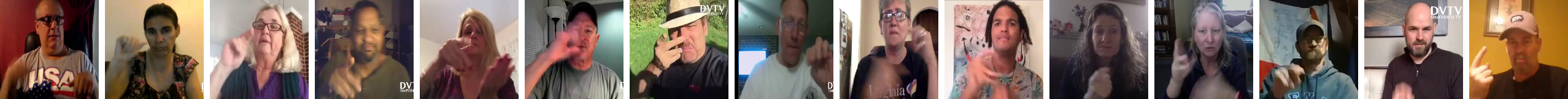}}\\
\midrule    
    misc & \raisebox{-.3\height}{\includegraphics[width=0.8\linewidth]{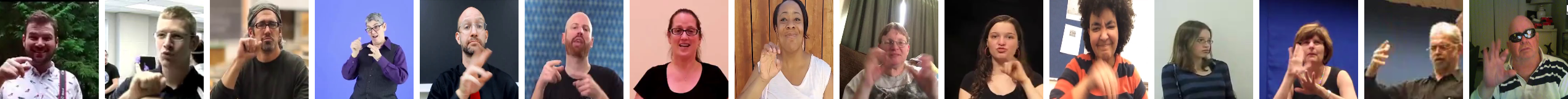}} \\
\bottomrule
\end{tabular}
\end{figure*}

 Table~\ref{tab:data-stats} shows the number of video clips in the two datasets. Figure~\ref{tab:data-stats} shows the distribution of fingerspelling sequence length in the two datasets. Figure~\ref{fig:source-example} shows image examples from the following three data sources:  YouTube, DeafVIDEO, misc.

\begin{table}[htp]
    \centering
    \caption{\label{tab:data-stats}\kledit{Numbers} of 300-frame video clips in ChicagoFSWild and ChicagoFSWild+.}
    \begin{tabular}{lrrr}
    \toprule
        Dataset & Train  & Dev & Test \\
        \midrule
        ChicagoFSWild & 3,539 & 691 & 673 \\
        ChicagoFSWild+ & 13,011 & 867 & 885 \\
        \bottomrule
    \end{tabular}
\end{table}

\begin{figure}[h]
    \centering
    \caption{Distribution of fingerspelling sequence length in ChicagoFSWild and ChicagoFSWild+}.
    \includegraphics[width=0.8\linewidth]{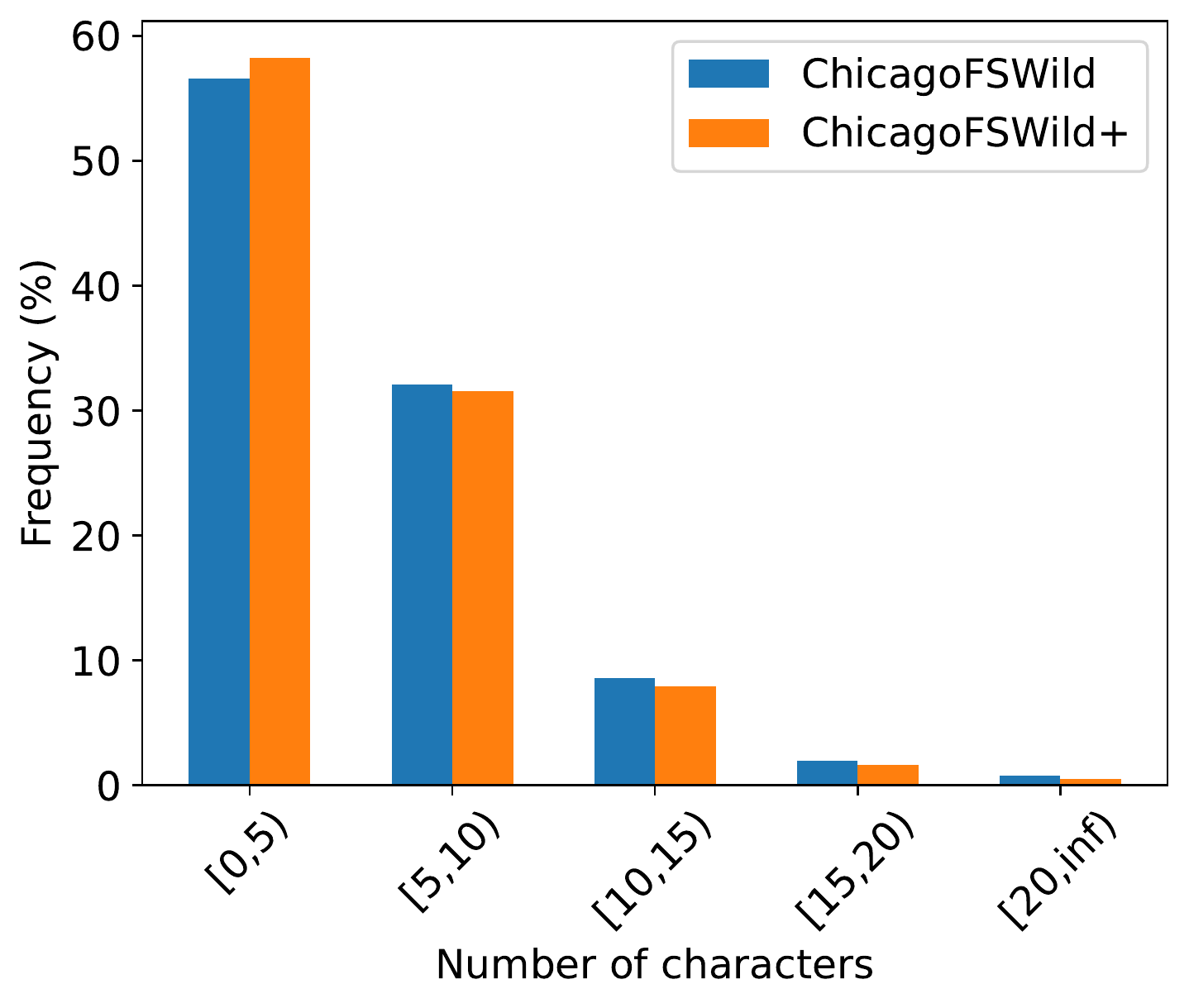}    \includegraphics[width=0.8\linewidth]{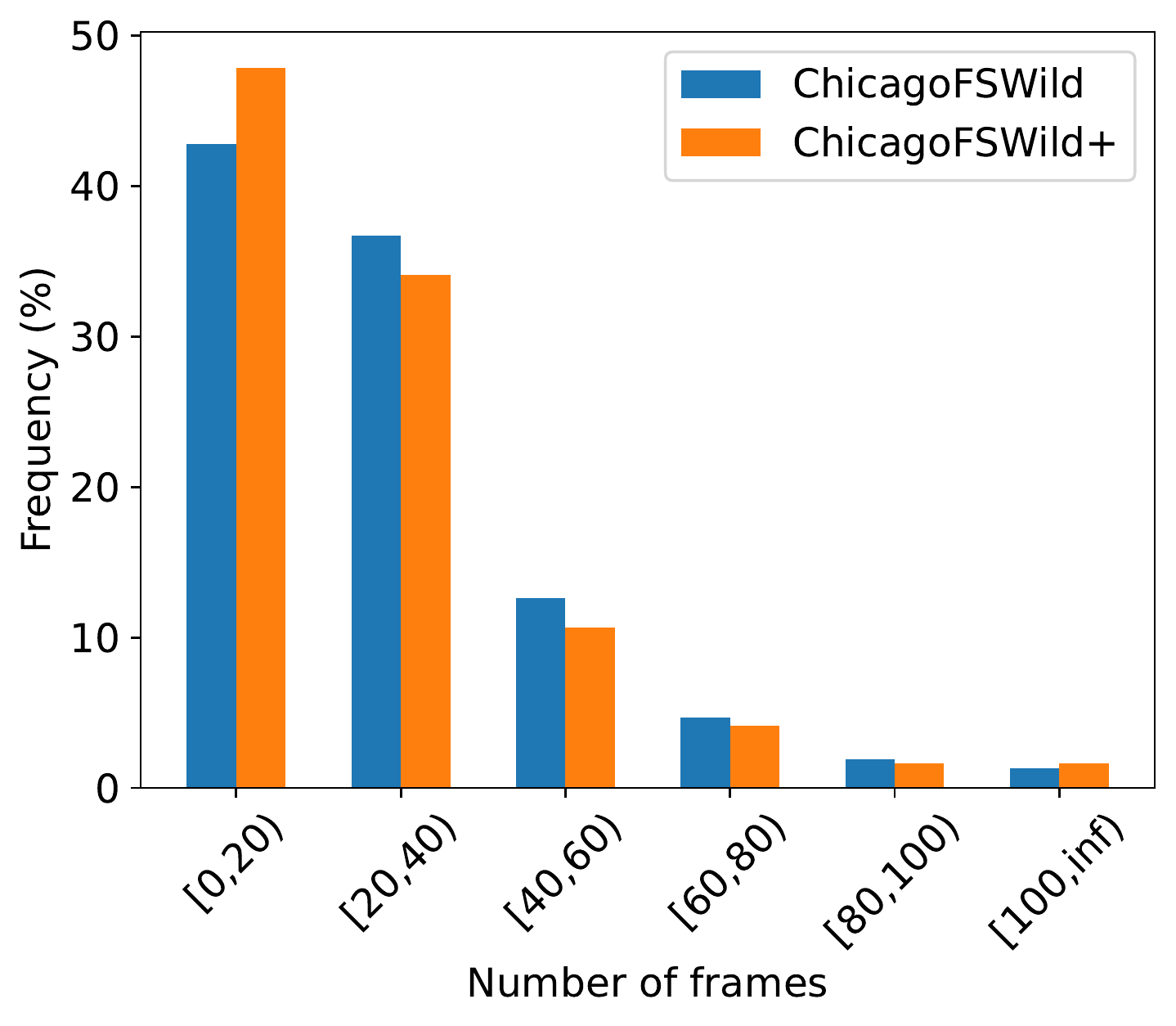}
\end{figure}

\subsection{Implementation Details}
\label{sec:app-impl}
\textbf{Pre-processing} The raw images in ChicagoFSWild and ChicagoFSWild+ datasets contain diverse visual scenes which can involve multiple persons. We adapt the heuristic approach used in~\citep{shi2019iccv} to select the target signer. Specifically, we use an off-the-shelf face detector to detect all the faces in the image. We extend each face bounding box by 1.5 times size of the bounding box in 4 directions and select the largest one with highest average magnitude of optical flow~\citep{farneback2003optical}. We further use the bounding box averaged over the whole sequence to crop the ROI area, which roughly denotes the signing region of a signer. Each image is resized to $160\times160$ before feeding into the model.

\textbf{Model implementation} The backbone convolutional layers are taken from VGG-19~\citep{Simonyan15vgg}. We pre-train the convolutional layers with a fingerspelling recognition task using the video-text pairs from the corresponding dataset. In pre-training, the VGG-19 layers are first pre-trained on ImageNet~\citep{imagenet} and the image features further go through a 1-layer Bi-LSTM with 512 hidden units per direction. The model is trained with CTC loss~\citep{ctc}. The output labels include the English alphabet plus the few special symbols, <space>, ', \&, ., @, as well as the blank symbol for CTC. The model is trained with SGD with batch size 1 at the initial learning rate of 0.01. The model is trained for 30 epochs and the learning rate is decayed to 0.001 after 20 epochs. The recognizer achieves 52.5\%/64.4\% lettter accuracy on ChicagoFSWild/ChicagoFSWild+ test sets. The VGG-19 convolutional layers are frozen in FSS-Net training.

In FSS-Net, the visual features output from convolutional layers are passed through a 1-layer Bi-LSTM with 256 hidden units per direction to capture temporal information. To generate proposals, we first transform the feature sequence via a 1D-CNN with the following architecture: conv layer (512 output dimension, kernel width 8), max pooling (kernel width 8, stride 4), conv layer (256 output dimension, kernel width 3) and conv layer (256 output dimension, kernel width 3). The scale of anchors is chosen from the range: $\{1, 2, 3, 4, 5, 6, 7, 8, 9, 10, 12, 14,$
$16, 18, 20, 24, 32, 40, 60, 75\}$,  according to the typical fingerspelling lengths in the two datasets. The positve/negative threshold of the anchors are 0.6/0.3 respectively. $\delta_{IoU}/\delta_{IS}$ are 1.0/0.8 (chosen from \{0.4, 0.6, 0.8, 1.0\}). The FS-encoder and text encoder are 3-layer/1-layer BiLSTM with 256 hidden units respectively. The margin $m$, number of negative samples in $\mathcal{N}_v$ and $\mathcal{N}_w$ are tuned to be 0.45, 5 and 5. The model is trained for 25 epochs with Adam~\citep{adam} at initial learning rate of 0.001 and batch size of 32. The learning rate is halved if the mean average precision on the dev set does not improve for 3 epochs. $\lambda_{det}$ in equation~\ref{eq:fss-overall-loss} is 0.1 (chosen from \{0.1, 0.5, 1.0\}). At test time, we generate $M=50$ proposals after NMS with IoU threshold of 0.7. $\beta$ is tuned to 1 (chosen from \{0.5, 1, 2, 3\}). 

\textbf{Implementation of Attn-KWS} The model assigns a score to video clip $\v I_{1:T}$ and word $w$ via equation~\ref{eq:attn-kws}, where $\v e_v^{1:T}$ is the visual feature sequence of $\v I_{1:T}$ and $\v e_x^w$ is the text feature of $w$, $\v W$ and $\v b$ are learnable parameters. The model is trained with cross-entropy loss.

\begin{equation}
\label{eq:attn-kws}
    \begin{split}
        & s(\v e_v^{t}, \v e_x^w)=\alpha(\frac{\v e_v^t\cdot\v e_x^w}{||\v e_v^t||\cdot||\v e_x^w||})^2+\theta\\
        & a(t)=\frac{\exp(s(\v e_v^{t}, \v e_x^w))}{\sum_t\exp(s(\v e_v^{t}, \v e_x^w))}\\
        & sc(\v I_{1:T}, \v e_x^w)=\sigma(\v W\displaystyle\sum_{t=1}^{T}a(t)\v e_v^{t} +\v b) \\
    \end{split}
\end{equation}

\subsection{Full results}
In addition to mAP and mF1, we also report ranking-based metrics: Precision@N and Recall@N (N=1, 10). For top-N retrieved X, we compute the percentage of correct X among N retrieved X as precision@N and the percentage of correct X among all correct X as recall@N, where X is text for FWS and video for FVS. Note the maximum value of R@1 and P@10 can be less than 1 as there are clips with multiple fingerspelling sequences and clips with fewer than 10 fingerspelling sequences. The performance of different models measured by all the above metrics is shown in table~\ref{tab:fss-perf-fswild-long}.

\begin{table*}[htb]
    \caption{\label{tab:fss-perf-fswild-long}FWS/FVS 
    \kledit{performance} on \kledit{the} ChicagoFSWild and ChicagoFSWild+ \kledit{test sets}. The maximum value of each metric is given in the parentheses (below each metric). \bsedit{The minimum value of each metric is 0.}
     }
    \centering
    \resizebox{2\columnwidth}{!}{
    \begin{tabular}{l c c c c c c  c c c c c c }
        \thickhline
        \multicolumn{1}{c}{} &
        \multicolumn{6}{c}{FWS (Video$\implies$Text)} &
        \multicolumn{6}{c}{FVS (Text$\implies$Video)}   \\
        \cmidrule(r){2-7}
        \cmidrule(r){8-13}
         \multicolumn{13}{c}{\textbf{ChicagoFSWild}} \\ 
        \cmidrule(r){2-7}
        \cmidrule(r){8-13}
        Method & mAP & mF1 & P$@$1 & P$@$10 & R$@$1 & R$@$10 & mAP & mF1 & P$@$1 &  P$@$10 & R$@$1 & R$@$10 \\
         & (1) & (1) & (1) & (.16) & (.75) & (1) & (1)  & (1)  & (1) &  (.17) & (.86) & (1)  \\
         \cmidrule(r){2-7}
         \cmidrule(r){8-13}
        Whole-clip & .175 & .154 & .116 & .043 & .092 & .293 & .142 & .119 & .106 & .039 & .070 & .251 \\
        Attn-KWS & .204 & .181 & .158 & .059 & .108 & .358 & .246 & .229 & .238 & .061 & .179  & .411\\
        Recognizer & .318 & .315 & .352 & .072 &.284 & .465 & .331 & .305 & .323 & .071 & .220 & .474  \\ 
        Ext-detector & .383 & .385 & .334  & .085 & .268 & .529 & .332 & .312 & .296 & .079 & .205 & .510 \\
        FSS-Net &\textbf{.434}&\textbf{.439}&\textbf{.384} & \textbf{.093} & \textbf{.300} & \textbf{.591} & \textbf{.394} & \textbf{.370} & \textbf{.370} & \textbf{.091} & \textbf{.255} & \textbf{.580}  \\ 
        \cmidrule(r){2-7}
        \cmidrule(r){8-13}
         \multicolumn{13}{c}{\textbf{ChicagoFSWild+}} \\ 
         \cmidrule(r){2-7}
        \cmidrule(r){8-13}
        Method & mAP & mF1 & P$@$1 & P$@$10 & R$@$1 & R$@$10 & mAP & mF1 & P$@$1 &  P$@$10 & R$@$1 & R$@$10 \\
        & (1) & (1) & (1) & (.16) & (.76) & (1) & (1)  & (1)  & (1) &  (.18) & (.84) & (1)  \\
        \cmidrule(r){2-7}\cmidrule(r){8-13}
        Whole-clip & .466 & .457&.416&.100&.326&.626
        & .548 & .526 & .546 & .101 & .421 & .711\\
        Attn-KWS & .545 & .530 & .485 & .112 & .392 &  .727 &
        .573 & .547 & .541 & .111 & .408 & .748 \\
        Recognizer & .465 & .462 & .470 & .094 & .390 & .620 & .398 & .405 & .394 & .090 & .292 & .617 \\
        Ext-detector &
        .633 & .641 & .589 & .118 & .491 & .769 &
        .593 & .577 & .568 & .114 & .419 & .786\\
        FSS-Net & \textbf{.674}
 & \textbf{.677} & \textbf{.637} &
\textbf{.123} & \textbf{.530} & \textbf{.796} & \textbf{.638} &
\textbf{.631} & \textbf{.596} &
\textbf{.123} & \textbf{.442} & \textbf{.825}\\ \thickhline
    \end{tabular}
    }
\end{table*}

\subsection{Examples of fingerspelling localization}
\begin{figure}[h]
\centering
\caption{\label{fig:localization}
\kledit{Examples of} fingerspelling localization produced by different methods. Upper: \textcolor{red}{Ground-truth}, Bottom: Attention \kledit{weight curve} and \textcolor{blue}{proposals} generated by our model. }
\begin{tabular}{c}
    \includegraphics[width=\linewidth]{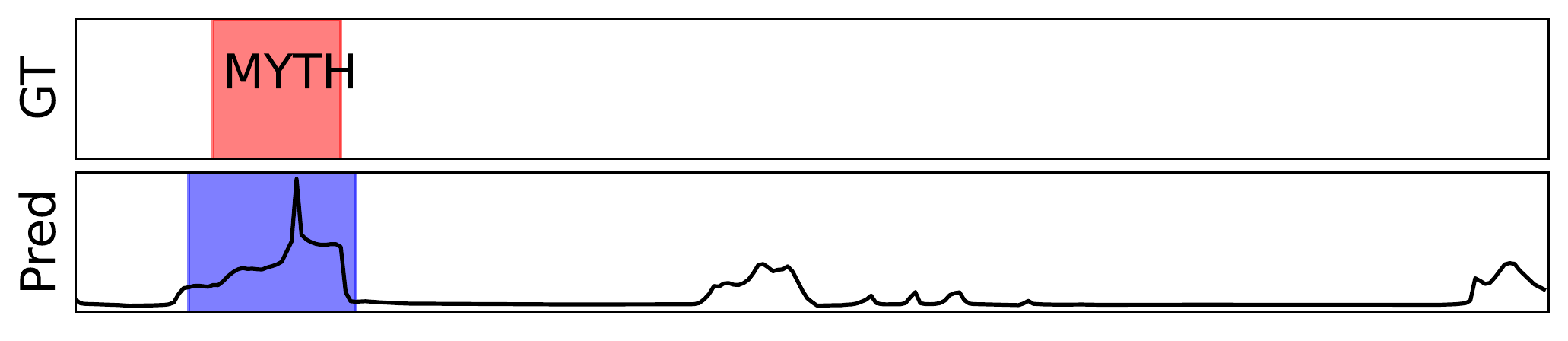}\\
    \includegraphics[width=\linewidth]{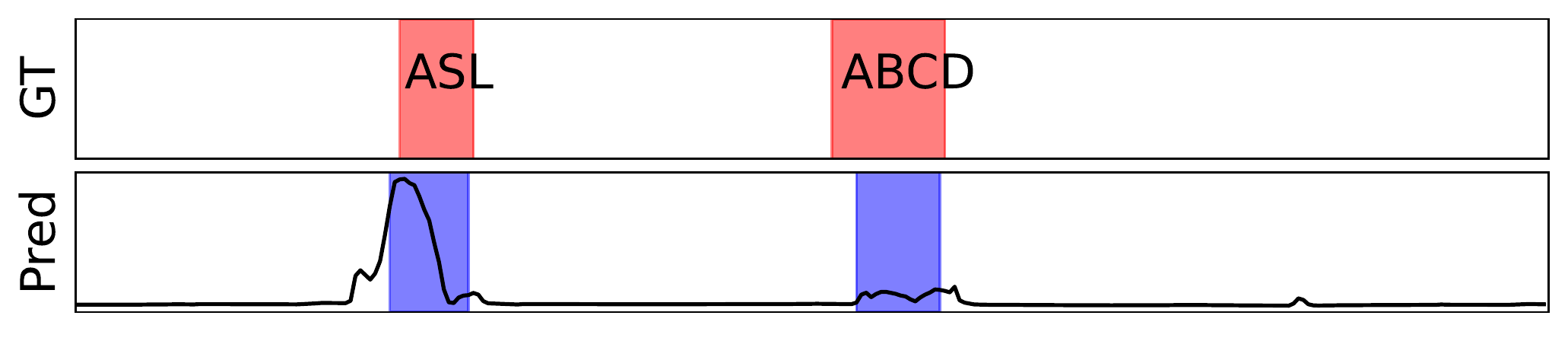} \\
    \includegraphics[width=\linewidth]{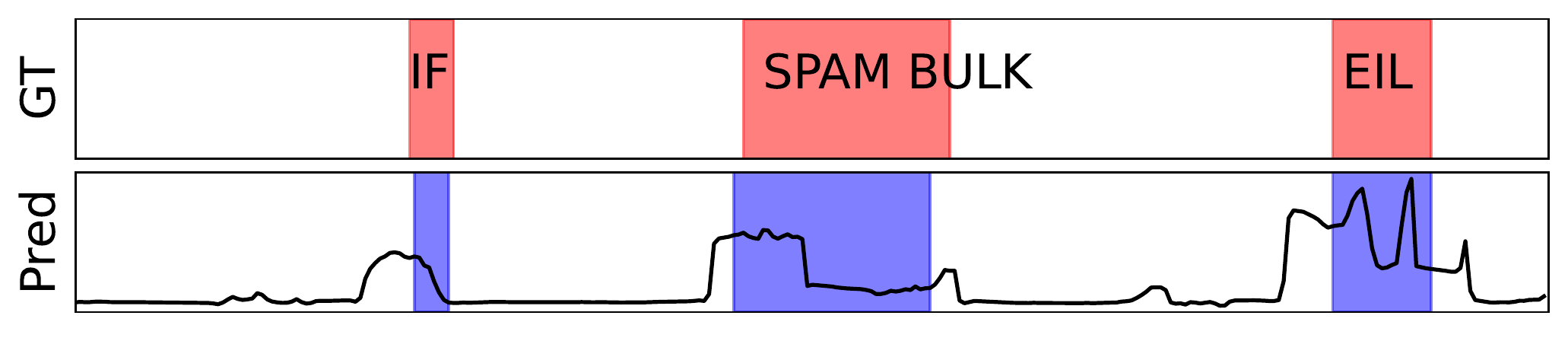}\\
    \includegraphics[width=\linewidth]{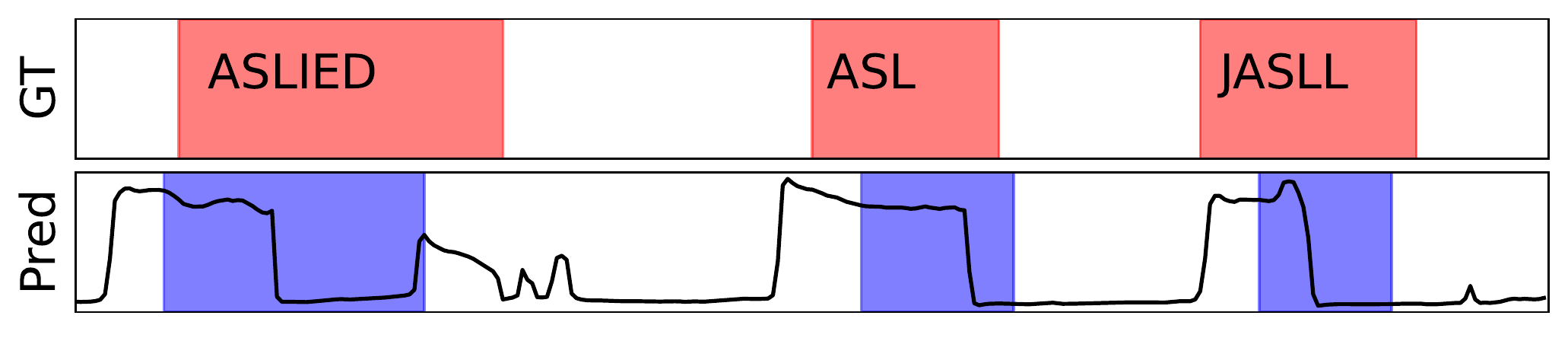} \\
    \end{tabular}
    
\end{figure}

Figure~\ref{fig:localization} shows examples fingerspelling localization produced by different methods.

\subsection{Precision-recall curve in FVS}
\label{sec:app-fvs-prcurve}
\begin{figure*}[htb]
\centering
\caption{\label{fig:pr-curves}FVS precision-recall curve of common words in ChicagoFSWild+ test set. Inside (): mAP}
\begin{tabular}{c@{}c@{}c}
    \includegraphics[width=0.32\linewidth]{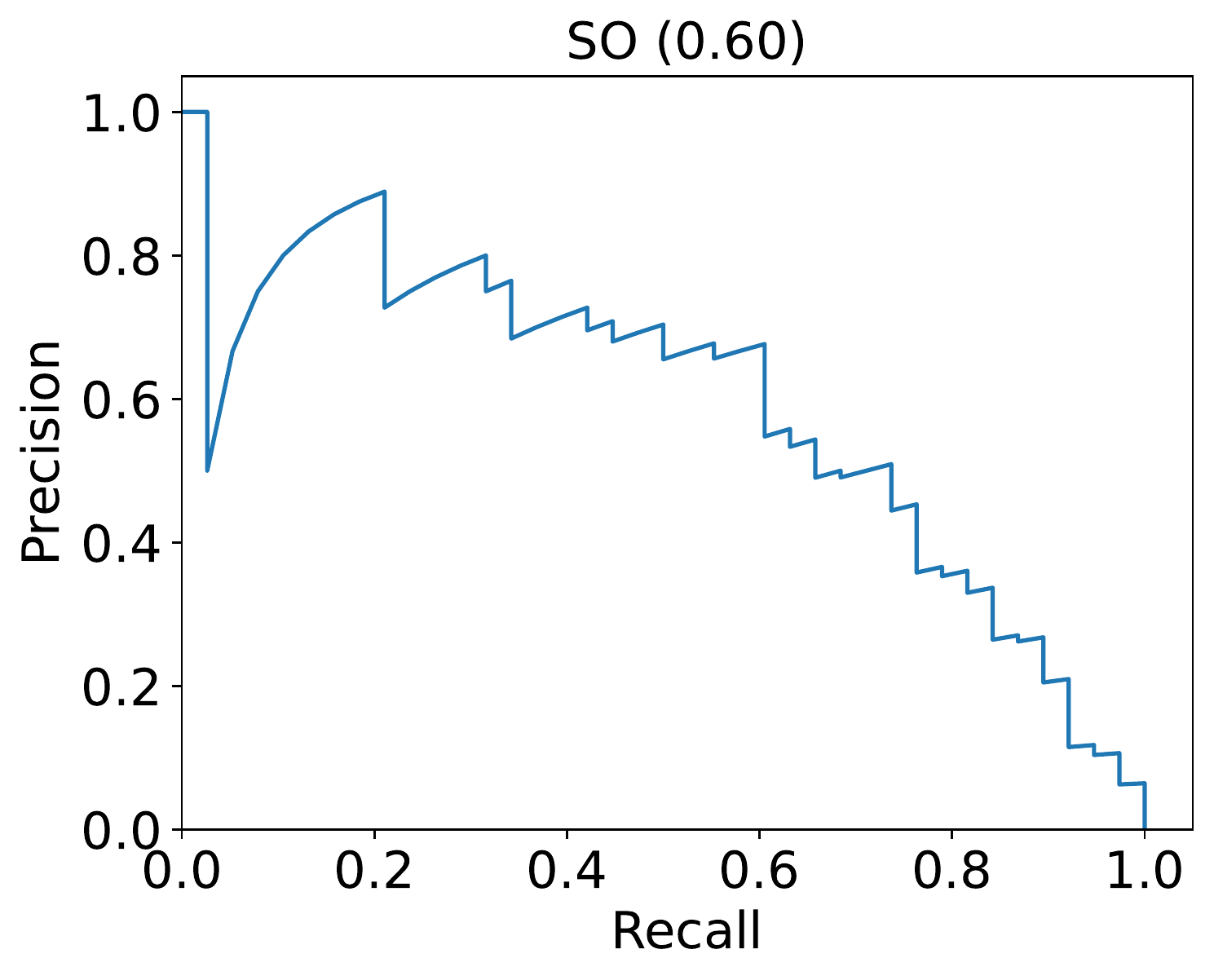} & \includegraphics[width=0.32\linewidth]{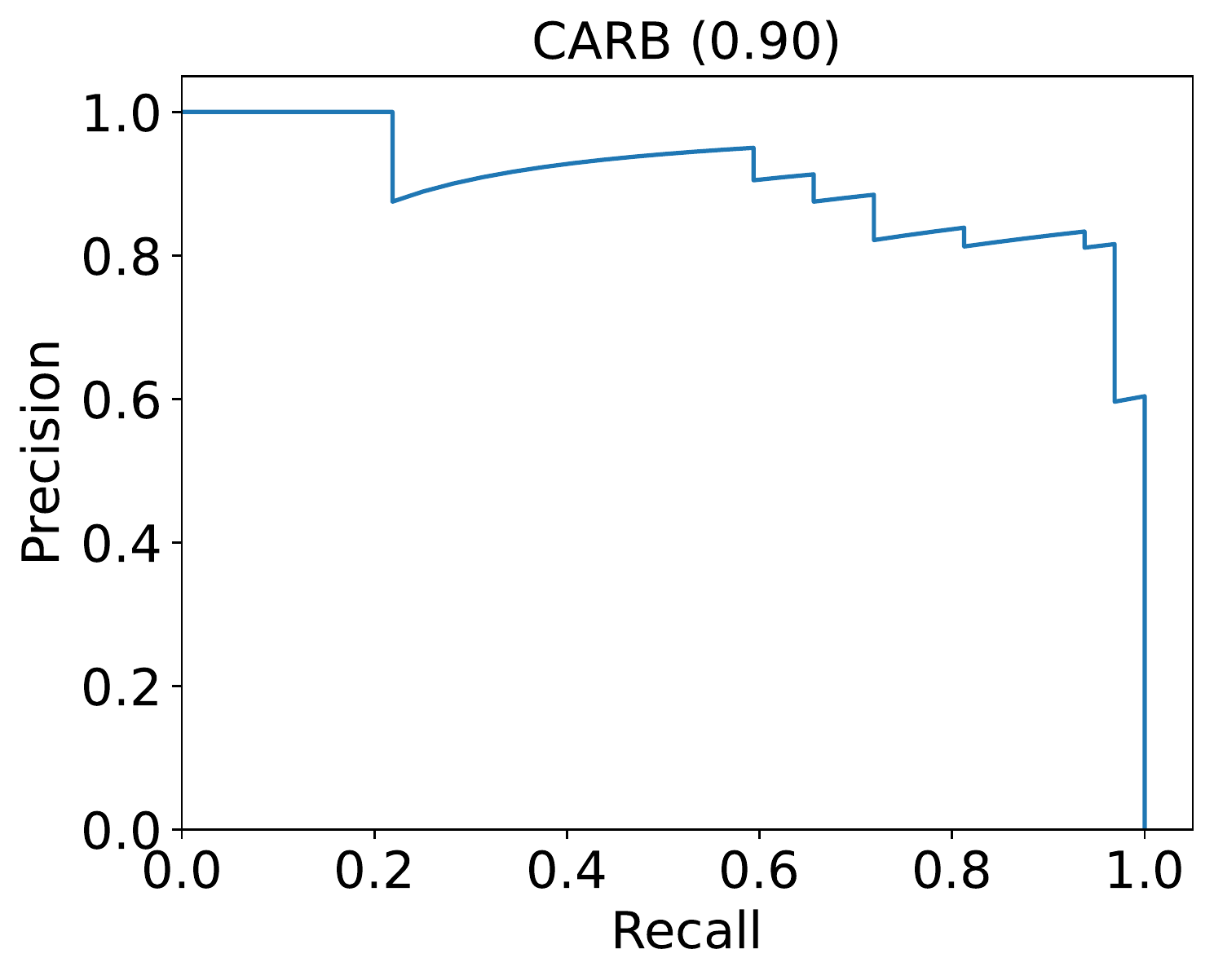} & \includegraphics[width=0.32\linewidth]{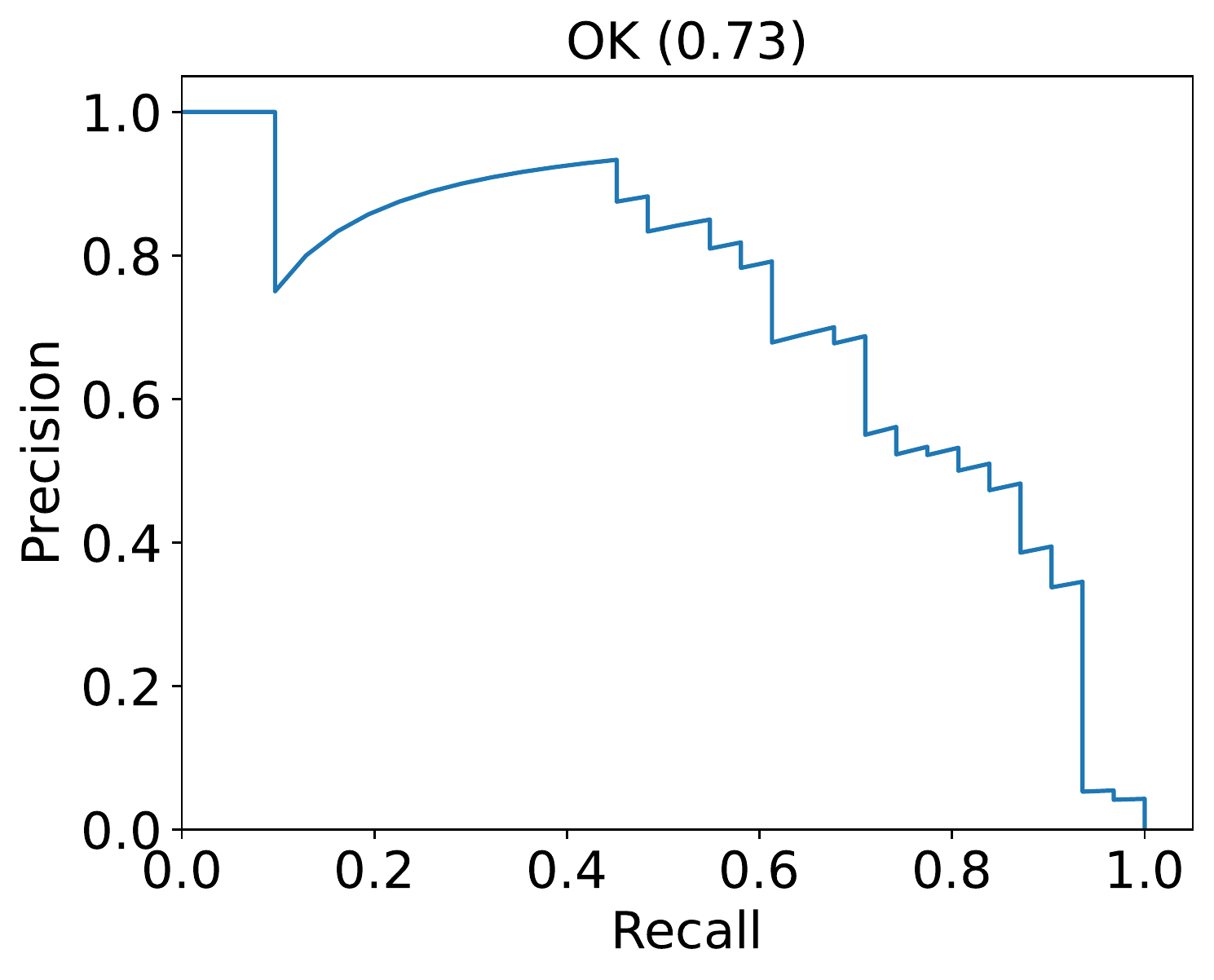} \\
    \includegraphics[width=0.32\linewidth]{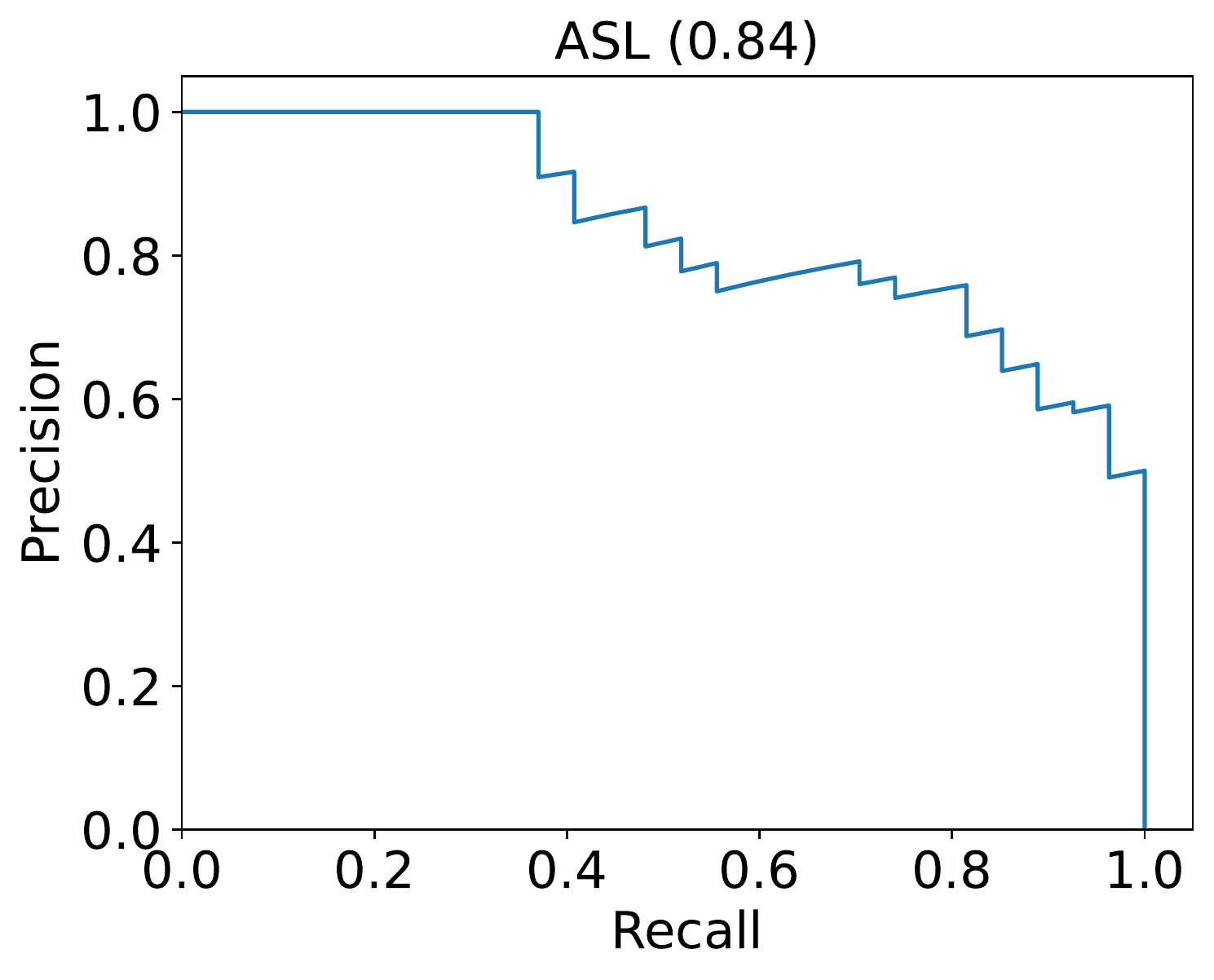} & \includegraphics[width=0.32\linewidth]{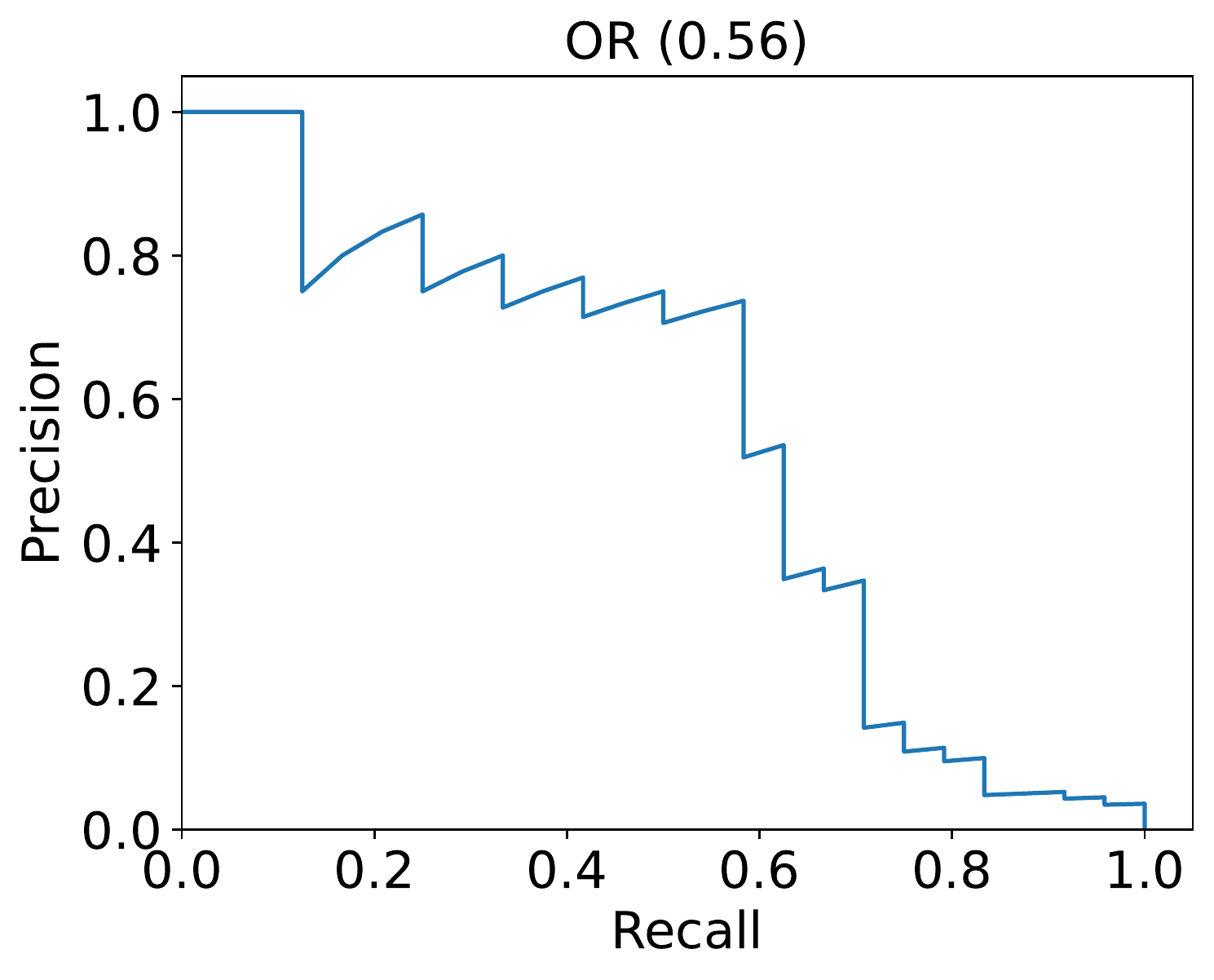} & \includegraphics[width=0.32\linewidth]{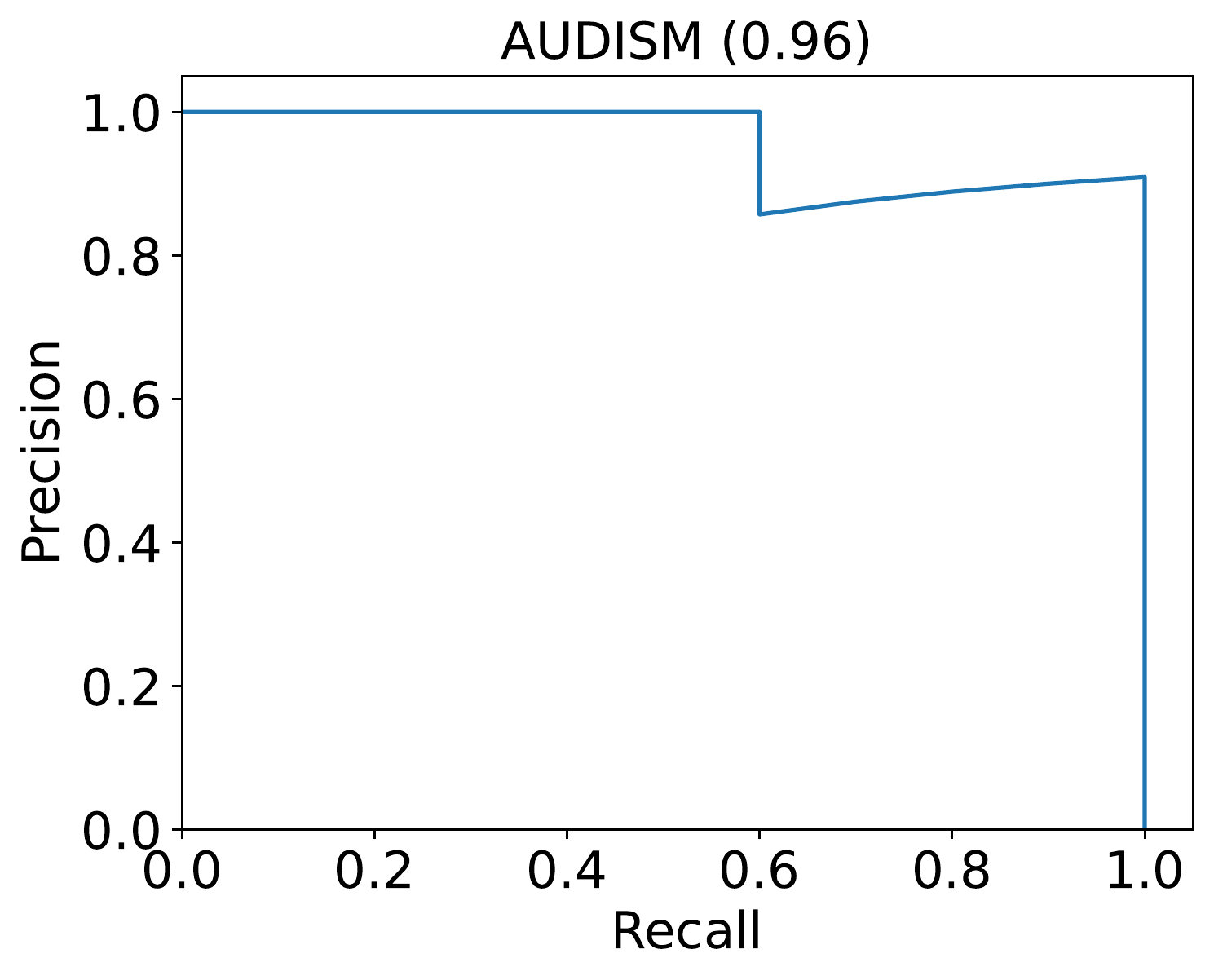} \\
    \includegraphics[width=0.32\linewidth]{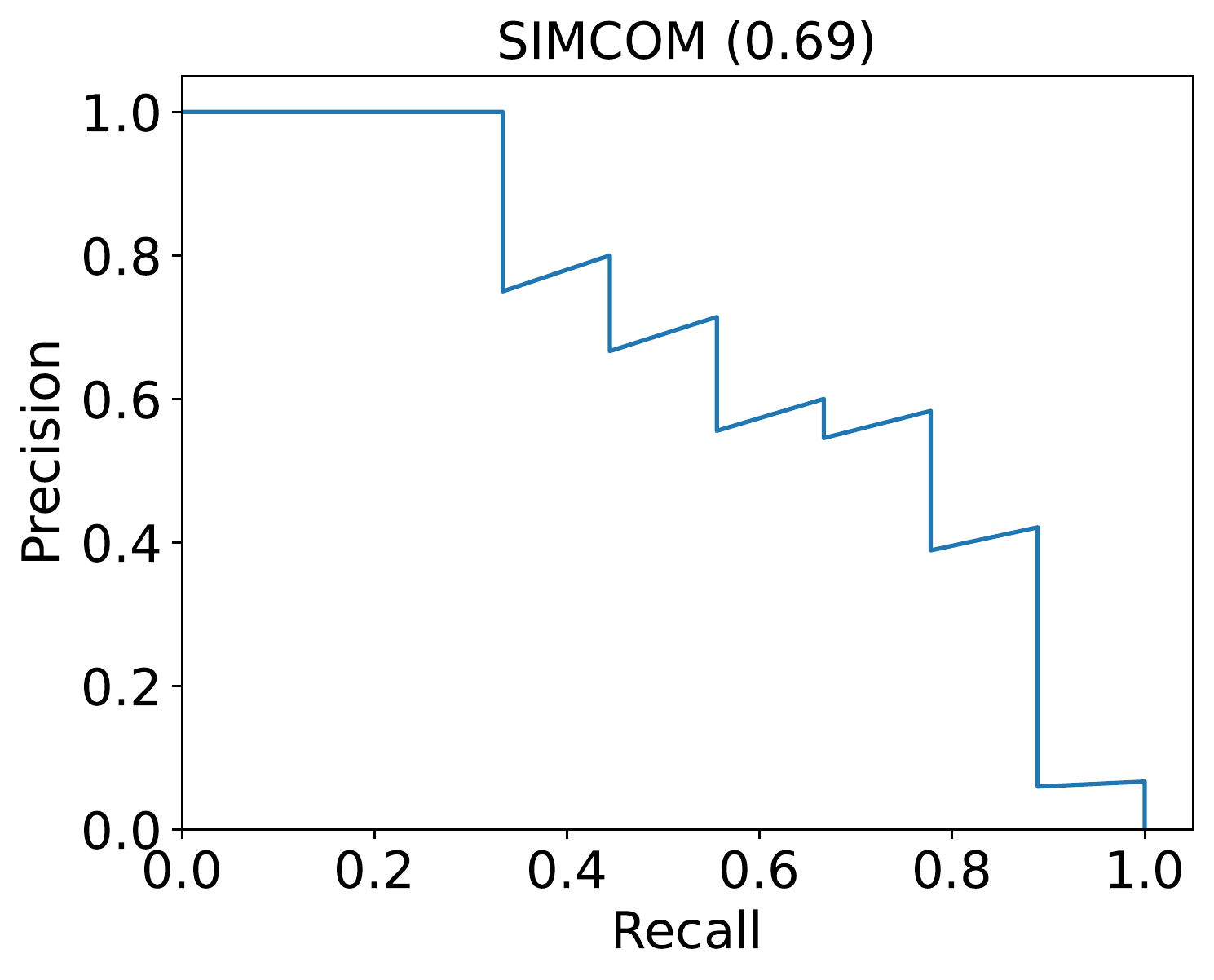} & \includegraphics[width=0.32\linewidth]{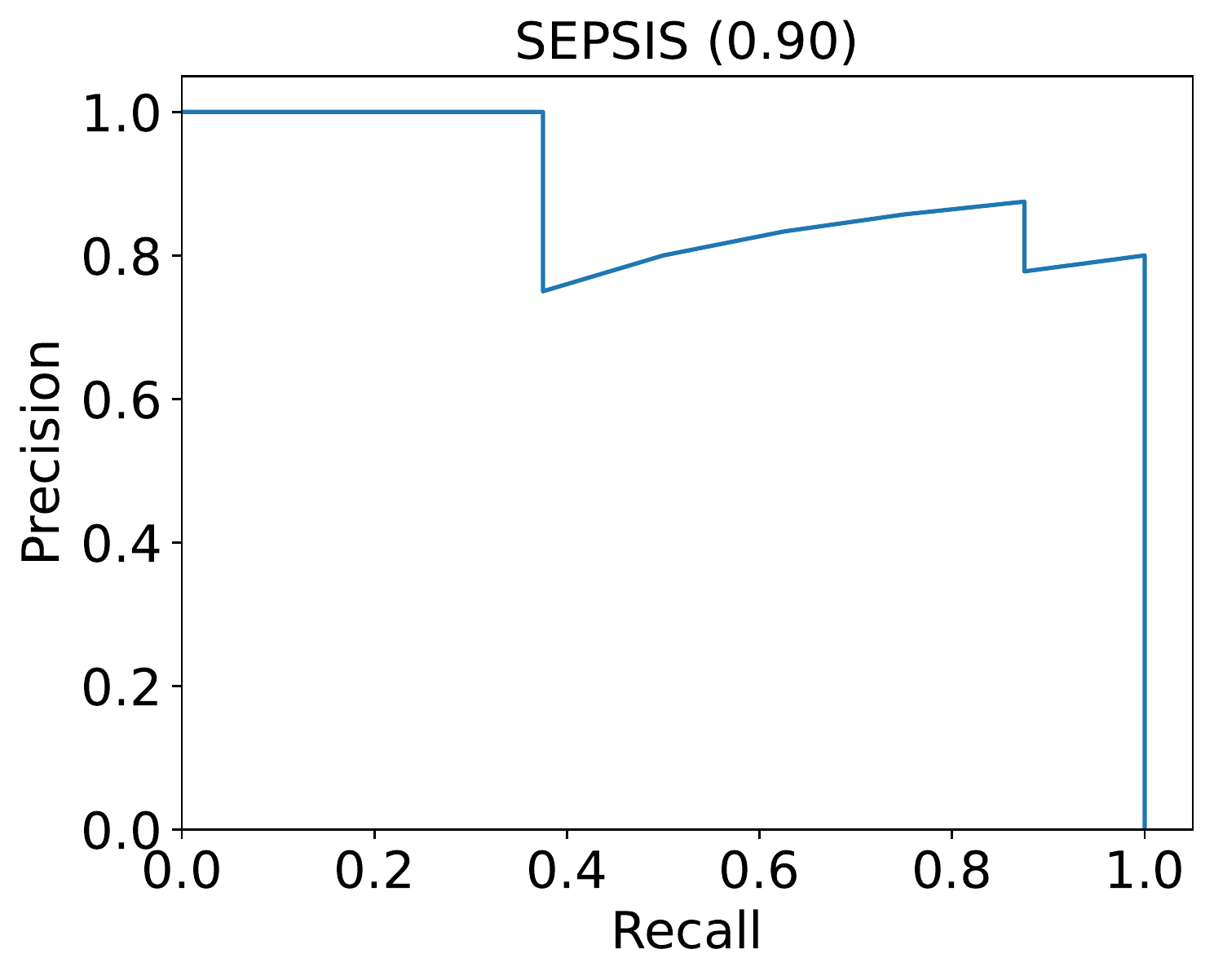} & \includegraphics[width=0.32\linewidth]{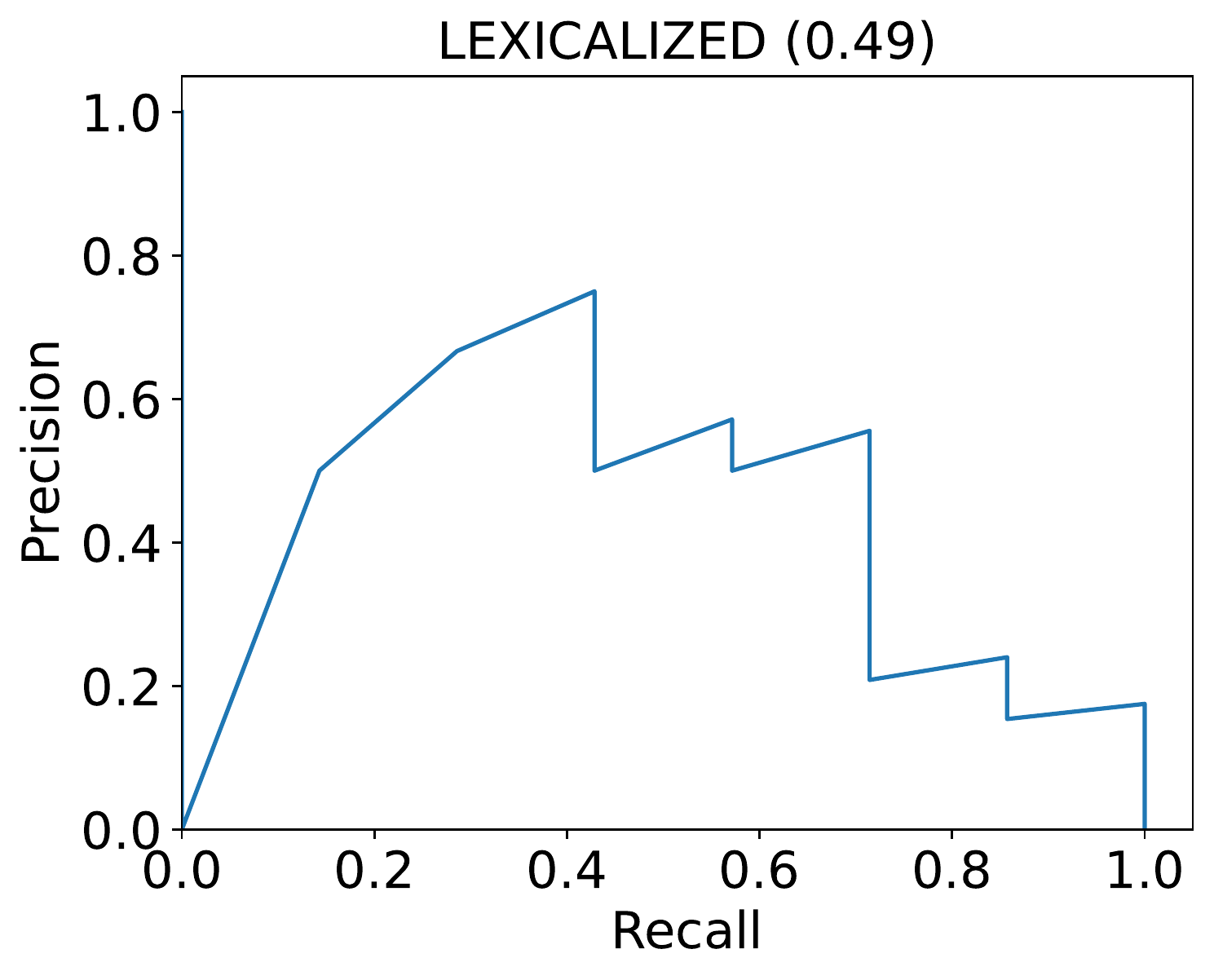} \\
    \end{tabular}
\end{figure*}

Figure~\ref{fig:pr-curves} shows the precision-recall curves of the most common words in the ChicagoFSWild+ test set. Overall the performance of our model on frequent words is higher than average. 

\subsection{Qualitative examples of pose estimation}
\label{sec:app-pose-rgb}
\begin{figure*}[htb]
\caption{\label{fig:openpose-examples} Estimated signer pose using OpenPose on the ChicagoFSWild test set.}
\includegraphics[width=\linewidth]{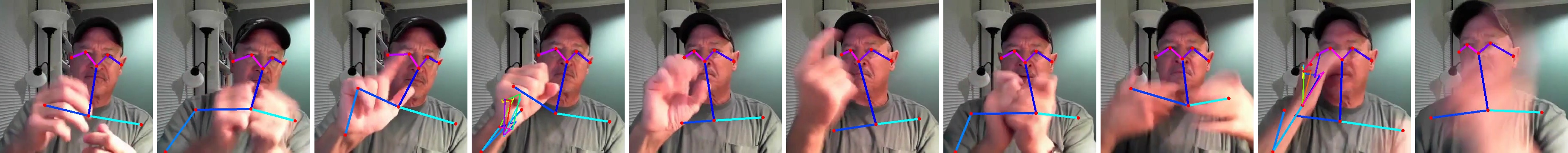}\\
\includegraphics[width=\linewidth]{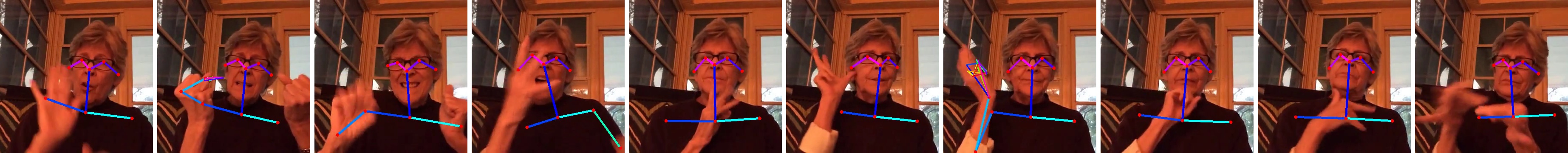}\\
\includegraphics[width=\linewidth]{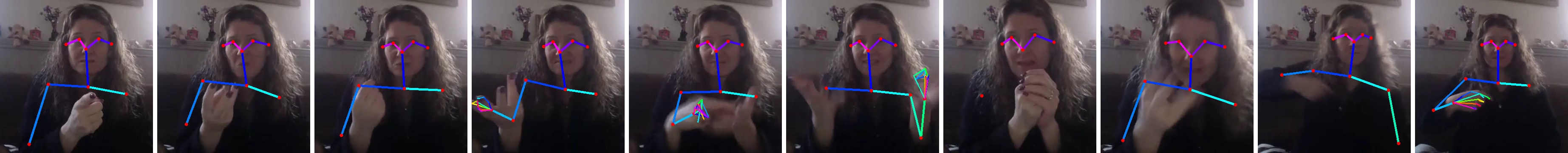}\\
\includegraphics[width=\linewidth]{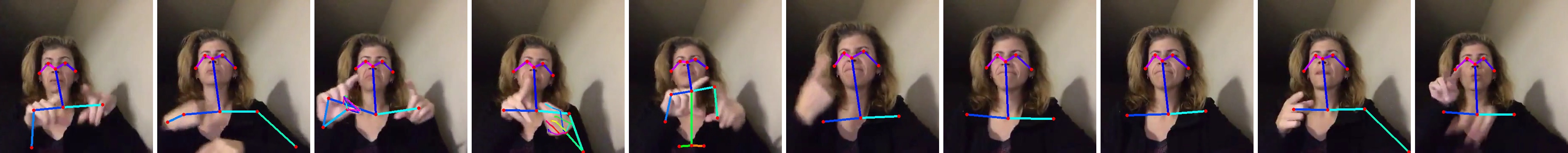}\\
\includegraphics[width=\linewidth]{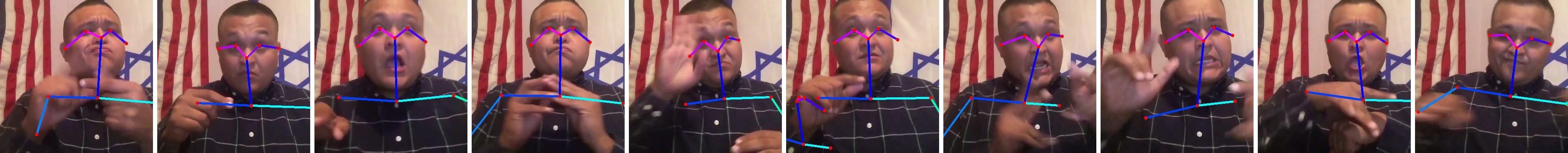}\\
\end{figure*}

Figure~\ref{fig:openpose-examples} shows typical failure cases of pose estimation on the ChicagoFSWild test set. The estimated hand pose is of low quality due to motion blur and hand occlusion.

\end{document}